\documentclass{article}

\usepackage[nonatbib, final]{neurips_2022}

\usepackage[utf8]{inputenc} 
\usepackage[T1]{fontenc}    
\usepackage{hyperref}       
\usepackage{url}            
\usepackage{booktabs}       
\usepackage{amsfonts}       
\usepackage{nicefrac}       
\usepackage{microtype}      
\usepackage{xcolor}         

\usepackage{amsmath}
\usepackage{amssymb}
\usepackage{mathtools}
\usepackage{amsthm}
\usepackage{wrapfig}
\usepackage{subcaption}

\theoremstyle{plain}
\newtheorem{theorem}{Theorem}[section]
\newtheorem{proposition}[theorem]{Proposition}

\theoremstyle{definition}

\theoremstyle{remark}

\title{Deep invariant networks with differentiable augmentation layers}

\author{%
    Cédric Rommel, Thomas Moreau \& Alexandre Gramfort\\
    Université Paris-Saclay, Inria, CEA, Palaiseau, 91120, France\\
    \texttt{\{firstname.lastname\}@inria.fr}
}

\usepackage{shortcuts}

\usepackage[textwidth=3cm, textsize=tiny]{todonotes}

\newcommand{\Cedric}[1]{\textcolor{black}{#1}}
\newcommand{\CedricRebuttal}[1]{\textcolor{black}{#1}}

\newcommand{\vset}{D_{\text{valid}}}
\newcommand{\tset}{D_{\text{train}}}
\newcommand{\loss}{\mathcal{L}}

\newcommand{\policy}{\mathcal{T}}

\begin{document}
\maketitle

\begin{abstract}
\looseness=-1
Designing learning systems which are invariant to certain data transformations is critical in machine learning.
Practitioners can typically enforce a desired invariance on the trained model through the choice of a network architecture, e.g. using convolutions for translations, or using data augmentation.
Yet, enforcing true invariance in the network can be difficult, and data invariances are not always known \emph{a piori}.
State-of-the-art methods for learning data augmentation policies require held-out data and are based on bilevel optimization problems,
which are complex to solve and often computationally demanding.
In this work we investigate new ways of learning invariances only from the training data.
Using learnable augmentation layers built directly in the network, we demonstrate that our method is very versatile.
It can incorporate any type of differentiable augmentation and be applied to a broad class of learning problems beyond computer vision.
We provide empirical evidence showing that our approach is easier and faster to train than modern automatic data augmentation techniques based on bilevel optimization, while achieving comparable results.
Experiments show that while the invariances transferred to a model through automatic data augmentation are limited by the model expressivity, the invariance yielded by our approach is insensitive to it by design.
\end{abstract}

\section{Introduction}

Inductive biases encoding known data symmetries are key to make deep learning models generalize in high-dimensional settings such as computer vision, speech processing and computational neuroscience, just to name a few.
Convolutional layers \cite{lecun1989backpropagation} are the perfect illustration of this, as their translation equivariant property allowed to reduce dramatically the size of the hypothesis space compared to fully-connected layers, opening the way for modern computer vision achievements \cite{krizhevsky_imagenet_2012}.
This illustrates one way to encode desired symmetries in deep learning models by hard-coding them directly in the network architecture.
An alternative way is to use data augmentation, where transformations encoding the desired symmetries are applied to the training examples, thus adding additional cost when such symmetry is not recognized by the model.
While in the first case invariances are \emph{built in} the network by design and are therefore a hard constraint, data augmentation promotes certain invariances more softly. As opposed to being  \emph{built-in}, desired invariances are here \emph{trained-in}~\cite{zhou_meta-learning_2020}.

In both cases, the invariances present in the data are not always known beforehand.
While relevant invariances are intuitive for some tasks such as object recognition (e.g. a slightly tilted or horizontally flipped picture of a mug, still represents a mug), the same cannot be said for many important predictive tasks such as classifying brain signals into different sleep stages \cite{chambon_deep_2017,xsleepnet}.
In order to be able to tackle this problem of learning optimal systems from complex data
two strategies are pursued in the literature:
neural architecture search (NAS) which aims to find the best architectural elements from the training data, and
automatic data augmentation (ADA) which aims to learn augmentation policies automatically from a given dataset.
Both fields tackle the problem very similarly,
by parametrizing the
network architecture or the
augmentation policies, leading to a bilevel hyperparameters optimization problem \cite{cubuk_autoaugment_2019}.
While these techniques allowed to find
architectures and
augmentations capable of outperforming the state-of-the-art in some cases, solving such bilevel optimization problems is often difficult and computationally demanding \cite{zela_understanding_2020, chen_stabilizing_2021, zhang_idarts_2021}.

This work investigates how to learn data invariances directly from the training data, avoiding the complex bilevel structure of ADA.
To this end, we propose to integrate learnable data augmentation layers within the network,
and train them together with other model parameters using a novel invariance promoting regularizer.
Our approach extends previous works into a very general framework that goes beyond computer vision tasks, as it can incorporate any type of differentiable augmentations.
We demonstrate the \emph{versatility} of our method on two well controlled simulated settings, as well as an image recognition and a sleep stage classification dataset.
We show that our approach can correctly select the transformations to which the data is invariant, and learn the true range of invariance, even for nonlinear operations.
Our experiments also demonstrate that the data augmentation layer proposed here leads to almost perfect built-in invariances irrespective to the complexity of the original network it is added to.
Moreover, we are able to achieve comparable performance and speed as state-of-the-art ADA approaches on our sleep staging experiment with a completely \emph{end-to-end} approach, avoiding tedious bilevel optimization parameters.
The accompanying code can be found at \url{https://github.com/cedricrommel/augnet}.

\section{Related work}

\paragraph{Automatic data augmentation}

Automatic data augmentation aims to learn relevant data invariances which increase generalization power.
More precisely, ADA is about searching augmentations that, when applied during the model training, will minimize its validation loss.
This objective is summarized in the following bilevel optimization problem:
\begin{equation}\label{eq:bilevel}
    \begin{aligned}
    \min_\policy  &\quad \loss(\theta^*|\vset)\\
    \text{s.t.} &\quad \theta^* \in \arg \min_{\theta} \loss(\theta|\policy(\tset)) \enspace ,
    \end{aligned}
\end{equation}
where $\policy$ is an augmentation policy, $\theta$ denotes the parameters of some predictive model, and $\loss(\theta|D)$ its loss over the set $D$.
Initial ADA approaches such as AutoAugment~\cite{cubuk_autoaugment_2019} and PBA~\cite{ho_population_2019} use discrete search algorithms to approximately solve \eqref{eq:bilevel}.
Despite the impressive results obtained, they are tremendously costly in computation time, which makes them impractical in many realistic settings.
In an attempt to alleviate this limitation, Fast AutoAugment~\cite{lim_fast_2019} proposes to solve a surrogate density matching problem, which breaks the bilevel structure of \eqref{eq:bilevel}.
It is hence substantially faster to solve, since it does not require to train the model multiple times.
However, this method needs a pre-trained model and its success highly depend on whether the latter was able to learn relevant data invariances on its own.
Another way of carrying ADA efficiently is by using gradient-based algorithms,
as proposed in Faster AutoAugment~\cite{hataya_faster_2020}, DADA~\cite{li_dada_2020} and ADDA~\cite{rommel2021cadda}.
While Faster AutoAugment also tackles a surrogate density matching problem, DADA and ADDA solve the bilevel problem \eqref{eq:bilevel} directly.
These ADA methods are the most related to our work since we rely on the same differentiable relaxations of standard augmentation transformations.
However, we are mostly interested in building the learned invariances into the model, which is out of the scope of these methods.
Indeed, they require substantially more overhead than our approach since they learn augmentations from the validation set and need to retrain the model after the augmentation search is completed.
Moreover, DADA and ADDA are based on an alternating optimization of the inner and outer problems~\cite{liu_darts_2019}, which suffer from noisy hypergradients on the outer level due to the stochastic inner problem, and require careful tuning of the outer learning rate \cite{zela_understanding_2020, chen_stabilizing_2021, zhang_idarts_2021}.

\paragraph{Embedding invariances within neural network architecture}

A vast literature has focused on encoding predefined invariances or equivariances into neural network architectures.
For instance, group convolutions allow to extend traditional convolutional layers to groups of affine transformations other than translations~\cite{cohenc2016group}.
More related to our work, DeepSets~\cite{Zaheer2017deep} encode permutation invariance by summing networks predictions.
Although related, these methods are not designed for learning symmetries from the data, which is the objective of this study.

Prior work on this matter from \cite{van_der_wilk_learning_2018}  proposes to learn invariances using the marginal likehood in the context of Gaussian processes.
In contrast, we are mostly focused on deep neural networks.
\cite{zhou_meta-learning_2020} suggests to discover data symmetries and build them into neural networks by learning weight sharing patterns in a meta-learning framework.
Most related to our work is Augerino~\cite{benton_learning_2020}, which allows to learn invariances to affine transformations from the training data, by adding a learnable sampling brick to an existing model and averaging its predictions.
They propose to learn the range of distributions from which affine transformations are sampled, by parametrizing the corresponding Lie group in terms of its Lie algebra.
As it is unclear how to extend Augerino beyond Lie groups such as affine transformations, its scope of application is rather limited, making it mostly taylored for computer vision tasks.
In this work we build on their ideas, extending Augerino to more diverse applications with hierarchical differentiable data augmentation layers built into the networks.
Moreover, in addition to learning sampling distribution ranges, we also learn to \emph{select} transformations which encode the strongest data invariance.

\section{AugNet: A general framework to embed data augmentation into neural networks}

\begin{figure*}[t]
    \centering
    \includegraphics[width=\linewidth, trim=0 3.5in 0 0, clip]{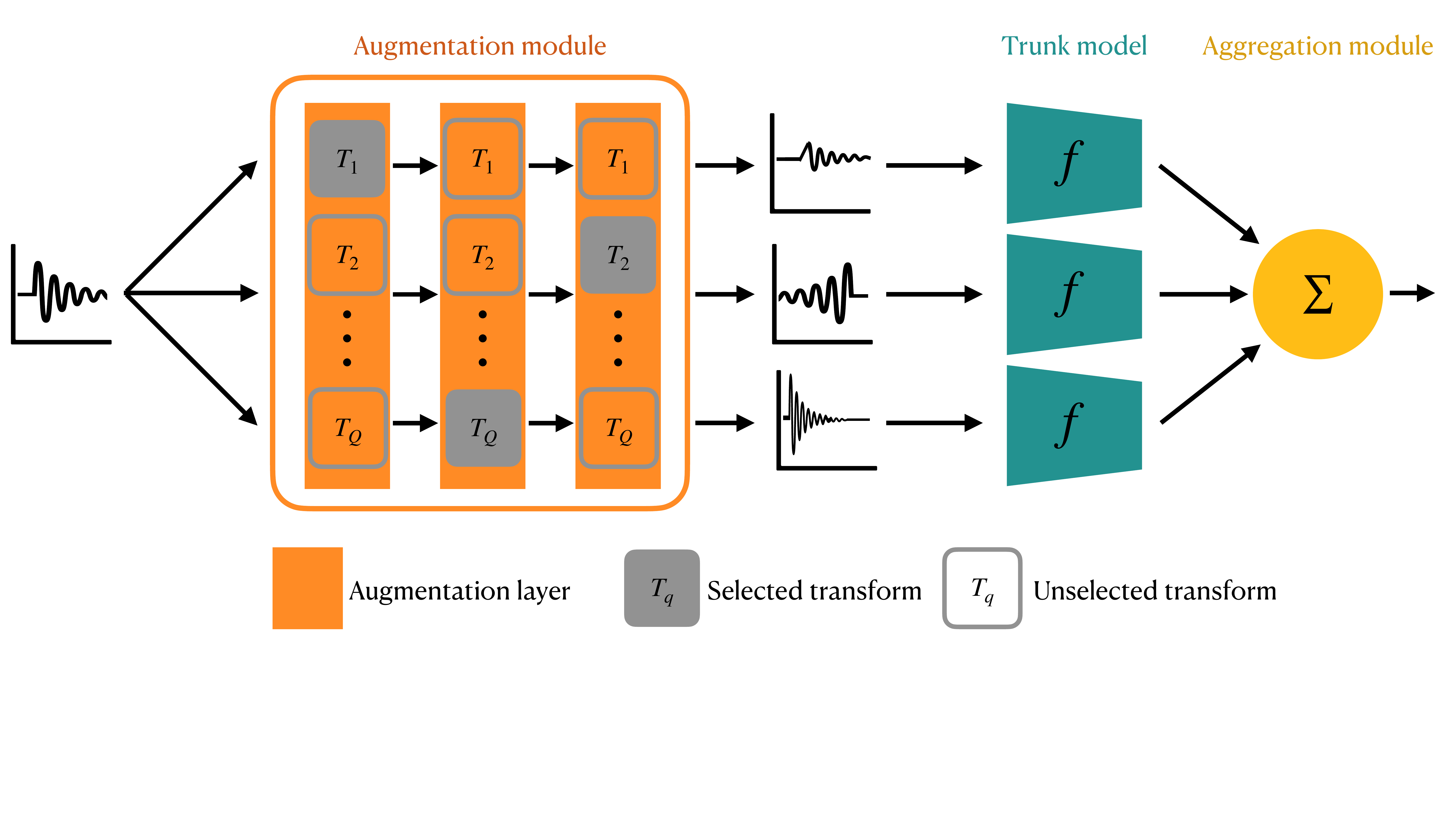}
    \caption{General architecture of AugNet. Input data is copied $C$ times and randomly augmented by the augmentation layers forming the \emph{augmentation module}. Each copy is then mapped by the \emph{trunk model} $f$, whose predictions are averaged by the \emph{aggregation module}. Parameters of both $f$ and the augmentation layers are learned together from the training set.}
    \label{fig:architecture}
\end{figure*}

\subsection{Preliminaries}

Let us consider a dataset $(x_1, y_1), \dots, (x_N, y_N)$ of observations sampled from an unknown distribution $\bbP_{X, Y}$ over $\cX \times \cY$.
In a supervised setting, one wants to use this data to train a model $f:\cX \to \cY$ (e.g. a neural network) so that it can predict $y$ from new observations $x$ sampled from the marginal $\bbP_X$.
Now, suppose that the data joint distribution $\bbP_{X, Y}$ is invariant to a certain group $G$ of transformations acting on $\cX$, i.e. for any $g\in G$ the distribution $\bbP_{X, Y}$ is close (in some sense) to the transformed distribution $\bbP_{gX, Y}$ (\lcf \cite{chen_group-theoretic_2020} for further details).
This means for example that, if $Y=y$ is the class "dog", than the probability of sampling an image $X=x$ of a dog is close to the probability of sampling the transformed image $gx$.
In this situation, one would like the model $f$ to have this same invariance by design, so that it does not have to learn it from scratch and can better generalize to new observations of $\bbP_{X, Y}$.
In other words, we would like $f(gx)=f(x)$ for any $g \in G$.
One way of achieving this is to average $f$ over $G$ endowed with a uniform distribution $\nu_G$:

\begin{proposition}
\label{eq:orbit-average}
For a given model $f:\cX \to \cY$ and a group of transformations $G$ acting on $\cX$, $\bar{f}: \cX\to\cY$ defined as
\[
    \bar{f}(x)=\bbE_{g \sim \nu_G}\left[f(gx) \right] \enspace ,
\]
where $\nu_G$ is a uniform distribution on $G$, is invariant through the action of $G$.
\end{proposition}

The proof follows from the assumption that $\nu_G$ is uniform and $G$ is a group, hence stable by composition and inverse. So for any $u \in G$, $gu^{-1}$ is also in $G$ and
\begin{equation*}
    \bar{f}(ux)=\bbE_{g \sim \nu_G}\left[f(gux) \right] =
    \bbE_{hu^{-1} \sim \nu_G}\left[f(hx) \right]
    = \bbE_{h \sim \nu_G}\left[f(hx) \right] = \bar{f}(x) \enspace,
\end{equation*}
where $h=gu$.
In practice, we will consider sets of transformations $G$ which do not necessarily form a group (e.g. rotations within a given range).
Because of this, $\bar{f}$ is only approximately invariant, although it can get very close to perfect invariance, as shown in our experiments from \autoref{sec:model-capacity}.

\subsection{Architecture: augment, forward and aggregate}

As the averaged model \eqref{eq:orbit-average} is intractable,
one can approximate it with an empirical average
\begin{equation*}
    \Tilde{f}(x) = \frac{1}{C}\sum_{c=1}^{C}f(g_c x),
\end{equation*}
where $g_c$'s are sampled from $\nu_G$.
In practice, using a large number $C$ of sampled transformations would be prohibitive as well.
Fortunately, even when $C$ is small, $\Tilde{f}$ is an unbiased estimator of $\bar{f}$.
Hence stochastic gradient descent can be used to train \emph{exactly} $\bar{f}$ by minimizing the loss of $\Tilde{f}$.

Based on these observations, we propose to create nearly invariant neural networks made of three blocks:
\begin{itemize}
    \item an \emph{augmentation module}, which takes an input $x \in \cX$, samples $C$ transformations from $G$, and outputs $C$ transformed copies of $x$;
    \item a \emph{trunk model} $f$, which can be any neural network mapping transformed inputs to predictions;
    \item and an \emph{aggregation module}, which is responsible for averaging the $C$ predictions.
\end{itemize}
This general architecture is illustrated on \autoref{fig:architecture} and referred to as \emph{AugNet} hereafter.
Note that unlike standard data augmentation, the transformations distributions is part of the model and kept at inference.

\subsection{Augmentation layers}

There are two missing elements from the architecture described in the previous section, namely the choice of the set of transformations $G$ and how transformations are sampled from it.
Indeed, the true invariances present in a dataset are often unknown, which is why we propose to learn both the set of transformations and a parametrized distribution used to sample from them.

In this paper, we are mostly interested in transformations $T:\cX \to \cX$ defining a data augmentation, such as random rotations in image recognition.
Most often, such operations can transform the data with more or less intensity depending on a parameter $\mu$, called \emph{magnitude} hereafter.
If we take the same example of random rotations, the magnitude can be the maximum angle by which we are allowed to rotate the images.
In order to have an homogeneous scale for all transformations considered, we assume without loss of generality that magnitudes lie in the interval $[0, 1]$, with $\mu=0$ being equivalent to the identity (i.e. no augmentation) and $\mu=1$ being the maximal transformation strength considered.
By using the reparametrization trick \cite{schulman_gradient_2015} or some other type of gradient estimator such as straight-through \cite{bengio_estimating_2013} or relax \cite{grathwohl_backpropagation_2018}, these transformations can be made differentiable with respect to $\mu$ as shown for example by \cite{hataya_faster_2020} for image augmentations and by \cite{rommel2021cadda} for augmentations of electroencephalography signals (EEG).
One can hence learn the right magnitude to use from the data by backpropagating gradients through $T$.

But the correct transformation $T$ describing a data invariance is also supposed to be unknown and needs to be learned in addition to its magnitude.
Let $\cT=\{T_1, \dots, T_Q\}$ be a discrete set of transformations possibly describing a data invariance.
We propose to use backpropagation to learn which transformation to pick from this set by using a layer consisting of a weighted average of all transformations:
\begin{equation} \label{eq:aug-layer}
    \text{AugLayer}(x; w, \mu) = \sum_{q=1}^{Q} w_q T_q(x; \mu_q),
\end{equation}
where the weights $w_q$ sum to 1.
In practice, we optimize some other hidden weights $w'$ which pass through a softmax activation $w=\sigma(w')$.
As explained in \autoref{sec:regularizarion} and demonstrated in our experiments, this layer architecture favors a single transformation describing a correct data invariance and tune its magnitude.

As the data might be invariant to more than one transformation, these layers can be stacked on top of each other, so that each one learns a different transformation (see \autoref{fig:architecture}).
This is justified by the stability property of invariance for composition.
Indeed, if a function $h:\cX \to \cY$ is invariant to actions $u$ and $g$, then it must be invariant to their composition: $h((u \circ g) x) = h(u(gx)) = h(gx) = h(x)$.
Hence, we build augmentation modules with sequences of augmentation layers \eqref{eq:aug-layer} and learn their parameters $w, \mu$ from the training set together with the parameters of the trunk model.


\subsection{Selective regularizer}
\label{sec:regularizarion}

As illustrated in our experiments from \autoref{sec:mario}, if we train \emph{AugNet} with a standard loss such as the cross-entropy, the model tends to find the most unconstrained model: the one with $\mu=0$, which never augments the data.
The same was described for Augerino \cite{benton_learning_2020}, which is why its authors proposed to add a regularization $R$ pushing towards broader distributions:
\begin{equation} \label{eq:learning-objective}
    \min_{w, \mu, \theta} \ell(\Tilde{f}(x; w, \mu, \theta), y) + \lambda R(\mu) \enspace .
\end{equation}
\begin{wrapfigure}[25]{r}{0.4\textwidth}
    \centering
    \includegraphics[width=0.95\linewidth]{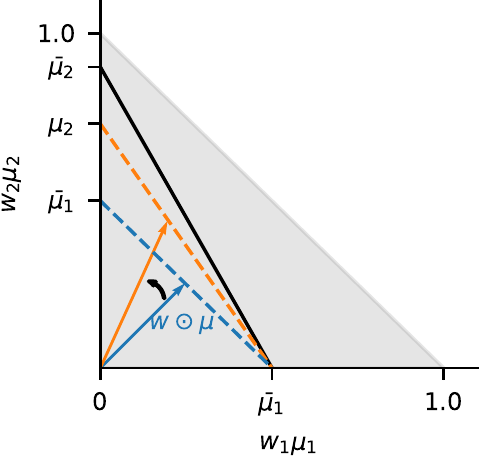}
    \caption{Illustration of the selective property of AugNet's regularizer.
    As the norm or the vector $w \odot \mu$ increases and $\mu_1$ reaches the maximal value allowed by the training loss $\bar{\mu}_1$,
    $\mu_2$ starts getting greater than $\mu_1$, creating an imbalance in the weights gradients.
    When $\mu_2$ reaches $\bar{\mu}_2$,
    the only way to keep increasing the vector's norm is by decreasing $w_1$ towards $0$ and increasing $w_2$ towards $1$.}
    \label{fig:reg-intuition}
\end{wrapfigure}
In the previous equation, $\theta$ denotes the parameters of the trunk neural network $f$ and $\ell$ is a loss function.
With our formalism, the regularization proposed by \cite{benton_learning_2020} is equivalent to penalizing the negative $L_2$-norm of the magnitude vector: $R(\mu)=-\| \mu \|_2$.
As illustrated in the ablation experiments of \autoref{sec:mario}, this regularizer is not sufficient to ensure that our augmentation layers do not converge to the identity transform because we have an additional degree of freedom than Augerino: the weights $w$.
Hence, the model may reach low loss values by just maximizing the weight of a single transform $w_q=1$ with magnitude $\mu_q=0$, while maximizing all other magnitudes $\mu_{q \neq q'} = 1$.
Because of this, we propose instead to penalize the norm of the element-wise product of weights and magnitudes: $R(w, \mu) = -\|w \odot \mu\|_2$.
This has the effect of tying together $w_q$ and $\mu_q$ of the same transformation $T_q$, avoiding the problem described before.

Another property of this regularization is that it promotes the selection of a single transformation per augmentation layer, as illustrated on \autoref{fig:reg-intuition} for the simple case of $n=2$ transformations.
Because $\sum_q w_q=w_1+w_2=1$, the vector $w \odot \mu=(x_1, x_2) \in [0, 1]^2$ is bound to move within the line of equation $x_2=\mu_2 (1 - \frac{x_1}{\mu_1})$ when the magnitudes $\mu_1$ and $\mu_2$ are fixed.
At the beginning of the training, with all weights initialized at $0.5$ and magnitudes close to $0$, if neither $T_1$ or $T_2$ harm the training, the vector's norm can grow following the bissector by increasing the magnitudes equally.
Once one of the transformations $T_1$ reaches a magnitude $\bar{\mu}_1$, it will start harming the training loss and its gradient will converge to 0.
This introduces an imbalance between the transformations as $\mu_2$ keeps increasing, until it reaches its maximal value $\bar{\mu}_2>\bar{\mu}_1$.
As $w \odot \mu$ is still bound to move within the line $\bar{\mu}_2 (1 - \frac{x_1}{\bar{\mu}_1})$, the only way to keep increasing its norm is by pointing more and more vertically, until it gets to $w_2=1$ and $w_1=0$.
This can also be seen from the expression of the gradients of the regularizer:
\begin{equation*}
    \frac{\partial R^2}{\partial w_i}(w, \bar{\mu}) = 2 \bar{\mu}_i^2 - 2 w_i (\bar{\mu}_1^2 + \bar{\mu}_2^2) \enspace .
\end{equation*}
which implies that:
\begin{equation*}
    \nabla_{w_1} R^2(w, \bar{\mu}) \leq \nabla_{w_2} R^2(w, \bar{\mu})
    \Longleftrightarrow
    \bar{\mu}_1 \leq \bar{\mu}_2 \enspace .
\end{equation*}
Each augmentation layer hence promotes the transformation with the highest admissible magnitude.

\section{Experiments with synthetic data}

In this section we present experimental results of \emph{AugNet} in controlled settings in order to verify empirically some of its properties and compare it to \emph{Augerino} and standard data augmentations.

\subsection{Comparison to Augerino}
\label{sec:mario}
\paragraph{Learning the correct invariance}

\begin{figure}[t]
    \centering
    \begin{subfigure}[b]{0.47\linewidth}
    \begin{center}
        \includegraphics[width=\linewidth]{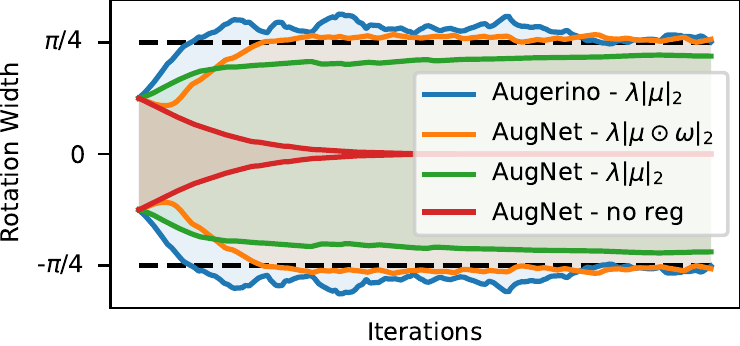}
        \caption{Mario\&Iggy experiment}
        \label{fig:mario-invariances}
    \end{center}
    \end{subfigure}
    \hfill
    \begin{subfigure}[b]{0.47\linewidth}
    \begin{center}
        \includegraphics[width=\linewidth]{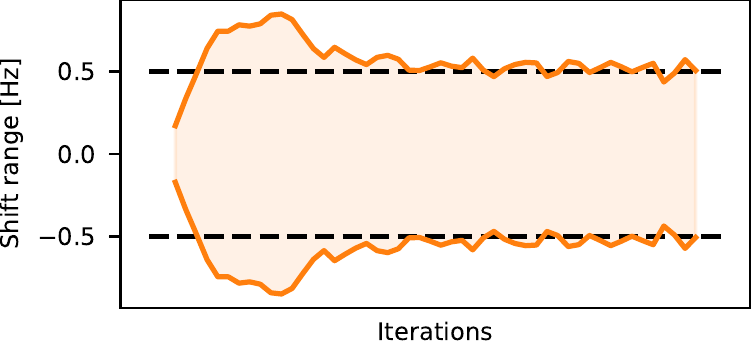}
        \caption{Sinusoidal waves experiment}
        \label{fig:learned-shift}
    \end{center}
    \end{subfigure}
    \caption{\textbf{(a)} Learned rotation angle during training. Both Augerino~\cite{benton_learning_2020} (blue) and AugNet (orange) are able to learn the correct level of rotation invariance in the data. When the regularization is removed from AugNet (red), it converges to the identity transform (no rotation) as observed in \cite{benton_learning_2020}. If we replace the regularizer in AugNet by Augerino's one, AugNet misses the correct rotation angle (green).
    \textbf{(b)} Learned frequency shift invariance. AugNet is able to learn nonlinear invariances from various types of data.
    }
\end{figure}

First, we reproduced the Mario-Iggy experiment by \cite{benton_learning_2020} to show that AugNet can also learn affine invariances from image datasets.
The data is generated from two initial images, which are rotated by a random angle either between $[-\pi/4, \pi/4]$ (labels 0 and 2) or between $[-\pi, -3\pi/4] \bigcup [3 \pi/4, \pi]$ (labels 1 and 3).
This procedure is illustrated in \autoref{fig:mario-data}.
Hence, labels depend both on the initial image and on whether it has its head pointing up or down.
By design, the data is invariant to inputs rotations between $[-\pi/4, \pi/4]$,
%
and the experiments objective
is to verify whether AugNet can learn this invariance.
For this, we use a single augmentation layer containing 5 geometric augmentations: \texttt{translate-x}, \texttt{translate-y}, \texttt{rotate}, \texttt{shear-x} and \texttt{shear-y}.
Note that while these augmentations are also encoded in Augerino, we don't use the Lie algebra and exponential maps to implement them here.
Maximal translations are set to half the image width/height, maximal rotations correspond to an angle of $\pm \pi$ and maximal shearing coefficients are $0.3$.
The trunk model used is a simple 5-layer convolutional network, whose architecture is described in \autoref{app:setting-mario} together with other experimental details.
As shown on
\autoref{fig:mario-invariances}, both Augerino (blue) and AugNet (orange) are capable of learning the correct angle of invariance from the data.

\paragraph{Ablation study}

\begin{figure*}[h]
    \centering
    \includegraphics[width=\linewidth]{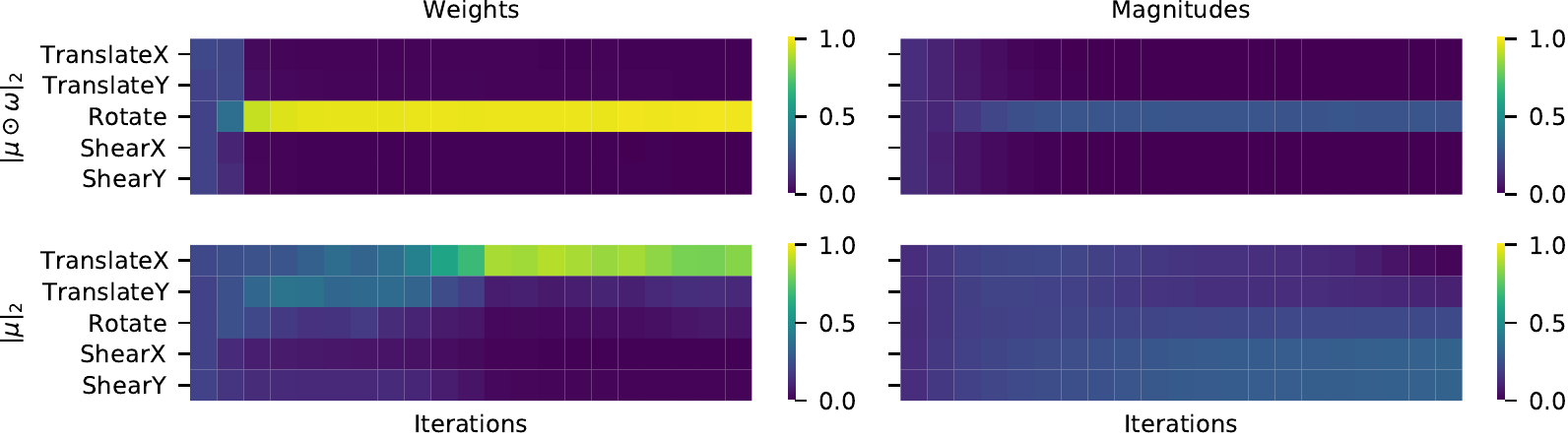}
    \caption{\textbf{Top:}
    With its standard regularizer,
    AugNet quickly learns to maximize the probability weight of rotation and adjusts the corresponding magnitude, ignoring other irrelevant transformations. \textbf{Bottom:}
    With Augerino's regularizer,
    the model may still converge to the identity map by maximizing the probability weight of any (irrelevant) transformation and minimizing its magnitude towards $0$.}
    \label{fig:mario-heatmap}
\end{figure*}

As explained in \autoref{sec:regularizarion}, the regularizer plays a crucial role both in AugNet and Augerino.
It can be seen on \autoref{fig:mario-invariances} that without regularization the rotation angle converges to $0$.
Indeed, our model naturally tends to nullify any transformations of the input,
similarly to what is
shown by \cite{benton_learning_2020} for Augerino.
Furthermore, we also see that when we replace AugNet's regularizer by Augerino's one ($-\|\mu\|_2$), the model does not learn the correct angle.
This happens because this penalty is not sufficient to prevent the model from converging to the identity transform, as shown by the learned weights and magnitudes depicted at the bottom of \autoref{fig:mario-heatmap}.
We see indeed that the model maximizes the weight for \texttt{translate-x} and minimizes its magnitude, while maximizing the magnitude of other transformations which have very small weights.
In contrast, we can see at the top of \autoref{fig:mario-heatmap} that the regularizer
we propose
allows to both \emph{select} the \texttt{rotate} operation and correctly tune its magnitude.

\subsection{Application to problems beyond computer vision}
\label{sec:freq-pred}
In this experiment we demonstrate that AugNet is also applicable to other types of data and can learn non-linear transformations which go beyond the scope of Augerino.
For this, we create a dataset consisting of four classes corresponding to 4 generating frequencies: $\omega =2, 4, 6$ and $8$ Hz.
For each of these frequencies, 10 seconds long sinusoidal waves of unit amplitude are created with frequencies sampled uniformly between $\omega \pm 0.5$ Hz.
Moreover, the phase of these waves are sampled uniformly and they are corrupted with additive Gaussian white noise with a standard-deviation of $0.5$ (\lcf \autoref{fig:sinusoids-dataset}).
We generate 400 training examples and 200 test examples.
The learning task consists in predicting the correct generating frequency $\omega$ from the noisy signals.
Note that, because classes are separated by $1$ Hz, the dataset presents an invariance to frequency shifts of the inputs between $\pm 0.5$ Hz.

In order to learn this invariance, we use a single augmentation layer implementing three augmentations from the EEG literature: \texttt{frequency~shift} \cite{rommel2021cadda}, \texttt{FT~surrogate} \cite{schwabedal_addressing_2019} and \texttt{Gaussian~noise} \cite{wang_data_2018}.
The trunk model used is a simple 3-layer convolutional network described in \autoref{app:setting-sinus} together with other experimental settings.
As shown in \autoref{fig:learned-shift}, AugNet is able to learn the correct range of invariance, and select the correct transformation.
\Cedric{This experiment was repeated in \autoref{app:more-sinusoids} with multiple augmentation layers to show that AugNet is robust to the chosen number of layers.}

\subsection{Insensitivity to model capacity}
\label{sec:model-capacity}
\looseness=-1
\begin{wrapfigure}[28]{r}{0.52\textwidth}
\vspace{-15pt}
\begin{center}
\includegraphics[width=\linewidth]{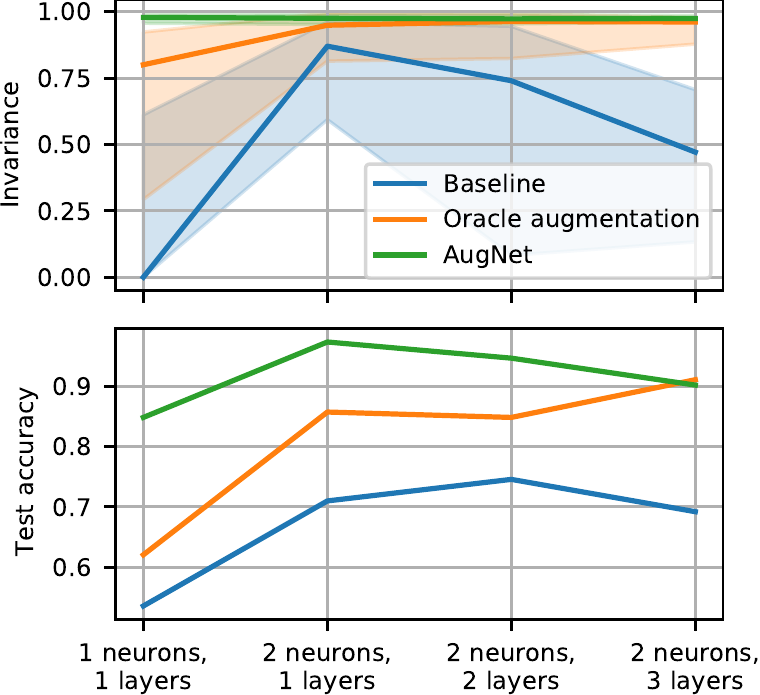}
\caption{\textbf{Top:} Model invariance \eqref{eq:inv} to the true frequency shift in the data across different architectures. We report the median values across the test set, with a 75\% confidence interval. Data augmentation helps to increase the model invariance, but is limited by its expressivity,
even when data invariances are known.
AugNet \emph{learns} to be almost perfectly invariant
and is insensitive to model capacity.
\textbf{Bottom:}
AugNet outperforms a model trained with \emph{oracle} augmentation with 3x fewer parameters.}
\label{fig:capacity-study}
\end{center}
\end{wrapfigure}
In this experiment we aim to demonstrate that models trained with our proposed methodology are invariant regardless of the capacity of the trunk model $f$ used, which represents an advantage over (automatic) data augmentation.
For this, we re-use the sinusoids dataset from \autoref{sec:freq-pred} 
and compare three methods: a baseline model trained directly on the raw data, the same model trained with an oracle augmentation $T$ (\texttt{frequency shift} with bounds of $\pm 0.5$ Hz) and an AugNet using the baseline as trunk model.
Unlike in \autoref{sec:freq-pred}, we use here a very small multi-layer perceptron with a varying number of neurons and layers, in order to be able to assess the impact of the model's expressivity.

In addition to the generalization power of the trained models, evaluated through their
test accuracy,
we are interested in how invariant they are to the true symmetry $T$ encoded in the data.
To gauge this property for a model $f$ at some input $x \in \cX$, we use the following metric adapted from \cite{bouchacourt2021grounding}:
\begin{equation} \label{eq:inv}
    \text{Inv}_T(f(x)) = \frac{b - d(f(x), f(T(x)))}{b} \enspace ,
\end{equation}
where $d$ is the cosine distance and $b=~d(f(x_i), f(x_j))$ is a baseline distance between randomly shuffled inputs
\footnote{Here $f$ refers to the activations before the final softmax.}.
As applying $T$ to the inputs of an invariant model should leave its outputs unchanged, the closer $\text{Inv}_T(f(x))$ is to $1$ for various inputs $x$, the more $f$ is invariant to $T$.
\looseness=-1

\autoref{fig:capacity-study} reports the median invariance across the test set of the three models for different number of layers and neurons.
It can be seen that data augmentation helps to increase the level of invariance of the baseline model considerably.
However, the level of invariance reached for the smallest models is lower than for larger models, suggesting that the former cannot encode perfectly the invariance taught to them through data augmentation.
This problem would be worse in a realistic setting where augmentations are more complex and imperfectly learned by ADA approaches rather than known \emph{a priori}.
On the contrary, results show that AugNet is capable of learning the correct augmentation from scratch and make the baseline model almost perfectly invariant to it, regardless of its expressivity.
Furthermore, the plot at the bottom of \autoref{fig:capacity-study} suggests that the level of invariance of each model is tightly related to its predictive performance.

\vspace{-0.3cm}
\section{Experiments with real datasets}

\vspace{-0.3cm}
\subsection{Image recognition}
\label{sec:cifar10}

%
In this experiment we showcase AugNet on a standard image recognition task using the CIFAR10 dataset~\cite{krizhevsky2009learning}.
All models considered in this experiment are trained for
\Cedric{300 epochs over 5 different seeds}
on a random 80\% fraction of the official CIFAR10 training set.
The remaining 20\% is used as a validation set for early-stopping and choosing hyperparameters, and the official test set is used for reporting performance metrics.
The trunk model used here is a pre-activated ResNet18~\cite{he2016identity, he2016deep}.

We consider
\CedricRebuttal{five}
baselines in this experiment: Augerino~\cite{benton_learning_2020},
\CedricRebuttal{AutoAugment~\cite{cubuk_autoaugment_2019} (pretrained for 5000h),
RandAugment~\cite{cubuk_randaugment_2020},}
as well as the trunk model trained with and without fixed data augmentations (\texttt{horizontal-flip} with probability 0.5 and \texttt{random-crop} of 32x32 pixels with a padding of 4 pixels~\cite{krizhevsky_imagenet_2012}).
\Cedric{For Augerino, AugNet and the baseline with fixed augmentations, we apply a standard normalization of RGB channels to the augmented examples before feeding them to the trunk model.}
%
%
\Cedric{AugNet's augmentation module is made of two layers containing the five augmentations detailed in \autoref{sec:mario} and a third layer with the following four different augmentations: \texttt{horizontal-flip}, \texttt{sample-pairing}~\cite{Inoue2018DataAB}, \texttt{contrast}, \texttt{brightness jitter}~\cite{Howard2014SomeIO}.}
Both AugNet and Augerino were trained with a number of copies set to $C=1$ and tested with
\CedricRebuttal{$C=20$ to take advantage of the invariant averaging architecture.}
The reader is referred to \autoref{app:setting-cifar} for further experimental details,
\CedricRebuttal{and to \autoref{app:more-cifar} for a sensitivity analysis on hyperparameters $C$ and $\lambda$.}

\begin{figure}[t]
    \centering
    \begin{subfigure}[b]{0.5\linewidth}
        \includegraphics[width=\linewidth]{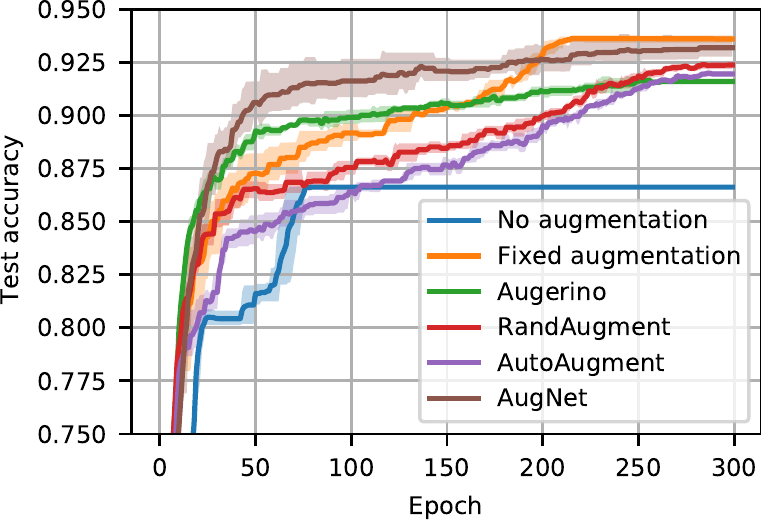}
        \caption{CIFAR10}
        \label{fig:cifar-benchmark}
    \end{subfigure}
    \hfill
    \begin{subfigure}[b]{0.44\linewidth}
        \includegraphics[width=\linewidth]{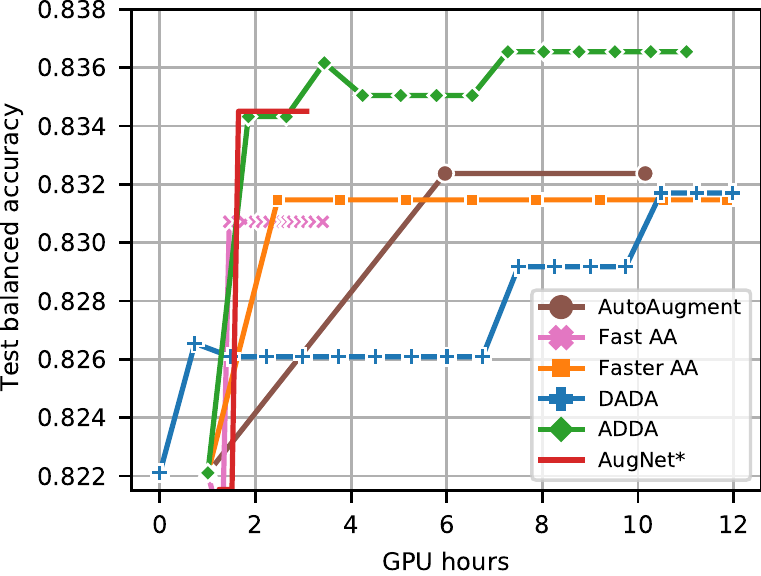}
        \caption{MASS}
        \label{fig:mass-benchmark}
    \end{subfigure}
    \caption{\textbf{(a)} Median test accuracy on CIFAR10, over 5 seeds, where
    shades represent 90\% confidence.
    AugNet learns invariances from scratch, outperforms automatic search approaches and achieves a performance comparable
    to state-of-the-art fixed data augmentations.
    \textbf{(b)} Median performance (over 5 folds) on the MASS dataset, as a function of the computation time.
    AugNet is comparable in speed and accuracy to ADDA, while being trained end-to-end. \CedricRebuttal{This figure is also available with error bars in the appendix \autoref{fig:mass-bars}}
    }
\end{figure}

%
\autoref{fig:cifar-benchmark} reports the performance across epochs.
For a given epoch, we report the test accuracy
corresponding to the best validation accuracy up to that point in training (equivalent to having early-stopped the model).
The performance of the baseline model with no augmentation
\CedricRebuttal{($86.6 \pm 0$\%)}
is considerably improved by
Augerino
\CedricRebuttal{($91.6 \pm 0.1$\%),
AutoAugment ($92.0 \pm 0.1$\%),
RandAugment ($92.4 \pm 0.2$\%)
}
and AugNet
\CedricRebuttal{($93.2 \pm 0.4$\%)}.
%
AugNet consistently outperforms
\CedricRebuttal{all three baselines}
over all seeds
\CedricRebuttal{within a 90\% confidence interval.}
This is probably due to AugNet selecting transformations which cannot be learned by Augerino, such as the \texttt{horizontal flip}
(\lcf \autoref{fig:cifar-selection-seed29}),
\CedricRebuttal{and to the better invariance of AugNet compared to models trained with data augmentation (\lcf \autoref{fig:cifar-inv}).}
%
\CedricRebuttal{Concerning the baseline with fixed augmentations ($93.6 \pm 0.2$\%), AugNet reaches a  comparable performance.
This demonstrates that in a realistic scenario with data from a real-world application where augmentations have not been explored as extensively as for CIFAR10, AugNet can be trained end-to-end and lead to a performance comparable to what we would get after manually trying many augmentation combinations.}
\looseness=-1

\vspace{-0.3cm}

\subsection{Sleep stage classification}
\label{sec:mass}

%
Similarly to \autoref{sec:freq-pred}, this experiment aims to demonstrate the practical usefulness of AugNet in a realistic setting beyond affine transformations.
For this, a sleep stage classification task is considered.
As most commonly done~\cite{Iber2007}, it consists in assigning to windows of 30\,s of EEG signals a label among five: Wake (W), Rapid Eye Movement (REM) and Non-REM of depth 1, 2 or 3 (N1, N2, N3).
The public dataset MASS - Session 3 \cite{o2014montreal} is used for this purpose (more details in \autoref{app:setting-mass}).
As done by \cite{rommel2021cadda}, both the training and validation sets consist of 24 nights each, and the test set contained 12 nights.
The trunk model used for this experiment is the convolutional network proposed in~\cite{chambon_deep_2017}, whose architecture is detailed in section \autoref{app:setting-mass} together with other experimental settings.
In this experiment, we compare the proposed approach to ADA methods.
For this, two discrete search methods were considered: \emph{AutoAugment}~\cite{cubuk_autoaugment_2019} and \emph{Fast AutoAugment}~\cite{lim_fast_2019}.
Additionally, three gradient-based methods were also tested: \emph{Faster AutoAugment}~\cite{hataya_faster_2020}, \emph{DADA}~\cite{li_dada_2020} and \emph{ADDA}~\cite{rommel2021cadda}.
All these methods shared the same policy architecture, consisting in 5 subpolicies made of 2 augmentation transformations
(\lcf \autoref{app:setting-mass}).
As for the previous experiment, we used two types of augmentation layers to implement AugNet: two layers capable of sampling \emph{time-frequency} transformations, and one layer made of \emph{sensors} transformations (\lcf \autoref{tab:mass-augmentations}).

%
\looseness=-1
\autoref{fig:mass-benchmark} presents the balanced accuracy over the test set as a function of the computation time.
For ADA methods,
we stopped the search every given number of steps (2 epochs or 5 samplings), used the learned policy to retrain the model from scratch (leading to a point in \autoref{fig:mass-benchmark}) and resumed the search from where it stopped.
For each run, we report the test accuracy of the best retrained model according to the validation accuracy.
Also note that some methods (Fast AutoAugment and ADDA) require a pretraining of the model.
Hence, the reported GPU hours for ADA methods correspond to the sum of pretraining, search and retraining time,
which explains why they start with some horizontal offset.
We see that, given a budget of 12 hours, AugNet is able to outperform four out of the five state-of-the-art approaches both in speed and accuracy.
%
%
It reaches a final performance comparable to the recently proposed ADDA, while requiring considerably less efforts in parameter fine tuning. AugNet only requires to set two hyperparameters: the regularization parameter $\lambda$ which is only one-dimensional, and the number augmentation layers stacked in the network. In contrast bilevel approaches require to carefully tune jointly the learning rate and the batch size of the validation set for the outer problem.
Moreover, while AugNet is trained \emph{end-to-end} once on the training set, ADDA requires pre-training the model and a final retraining of the model with the learned augmentation policy.
%
%

\vspace{-0.3cm}
\section*{Conclusion}
\vspace{-0.3cm}

In this paper we propose a new method coined AugNet to learn data invariances from the training data, thanks to differentiable data augmentation layers embedded in a deep neural network.
Our method can incorporate any type of differentiable augmentations and is applicable to a broad class of learning problems.
We show that our approach can correctly select the transformation to which the data is invariant, and learn the true range of invariance, even for nonlinear operations.
While (automatic) data augmentation is limited by the capacity of the model to encode symmetries, our approach leads to almost perfect invariance regardless of the model size.
On computer vision tasks, this advantage allows our method to reach high generalization power without hand-crafted data augmentations.
Promising results are also obtained for sleep stage classification, where AugNet outperforms most ADA approaches both in speed and final performance with \emph{end-to-end} training, avoiding tedious bilevel setup.
\looseness=-1
\vspace{-0.3cm}
\paragraph{Limitations}
\CedricRebuttal{In order to be able reach top performance, a large number of copies $C$ might be necessary, increasing AugNet's computational complexity at test time (\lcf \autoref{app:more-cifar}). This means that a proper trade-off between performance and speed has to be determined for each use-case.
Moreover, although AugNet is able to learn augmentations in a broader scope than previous end-to-end methods, it is still limited to augmentations that can be relaxed into differentiable surrogates and requires the availability of a set of possible augmentations $\mathcal{T}$. These limitations are not specific to AugNet however, as it is a standard requirement in ADA \cite{cubuk_autoaugment_2019, hataya_faster_2020, li_dada_2020, lim_fast_2019}.
}

\begin{ack}
This work was supported by the BrAIN grant (ANR-20-CHIA-0016) and ANR AI-Cog grants (ANR-20-IADJ-0002). It was also granted access to the HPC resources of IDRIS under the allocation 2021-AD011012284R1 and 2022-AD011011172R2 made by GENCI.
\end{ack}

{
\small
\bibliographystyle{unsrt}
\bibliography{augnet}

\begin{thebibliography}{10}

\bibitem{lecun1989backpropagation}
Y.~LeCun, B.~Boser, J.~S. Denker, D.~Henderson, R.~E. Howard, W.~Hubbard, and
  L.~D. Jackel.
\newblock Backpropagation applied to handwritten zip code recognition.
\newblock {\em Neural Computation}, 1:541--551, 1989.

\bibitem{krizhevsky_imagenet_2012}
Alex Krizhevsky, Ilya Sutskever, and Geoffrey~E. Hinton.
\newblock {ImageNet} classification with deep convolutional neural networks.
\newblock In {\em Advances in neural information processing systems (NeurIPS)},
  2012.

\bibitem{zhou_meta-learning_2020}
Allan Zhou, Tom Knowles, and Chelsea Finn.
\newblock Meta-{Learning} {Symmetries} by {Reparameterization}.
\newblock In {\em {International Conference on Learning Representations
  (ICLR)}}, 2021.

\bibitem{chambon_deep_2017}
Stanislas Chambon, Mathieu Galtier, Pierrick Arnal, Gilles Wainrib, and
  Alexandre Gramfort.
\newblock A deep learning architecture for temporal sleep stage classification
  using multivariate and multimodal time series.
\newblock {\em IEEE Transactions on Neural Systems and Rehabilitation
  Engineering}, 26(4):758--769, 2018.

\bibitem{xsleepnet}
Huy Phan, Oliver~Y. Chen, Minh~C. Tran, Philipp Koch, Alfred Mertins, and
  Maarten De~Vos.
\newblock Xsleepnet: Multi-view sequential model for automatic sleep staging.
\newblock {\em IEEE Transactions on Pattern Analysis and Machine Intelligence},
  2021.

\bibitem{cubuk_autoaugment_2019}
Ekin~D. Cubuk, Barret Zoph, Dandelion Mane, Vijay Vasudevan, and Quoc~V. Le.
\newblock {AutoAugment}: {Learning} {Augmentation} {Strategies} {From} {Data}.
\newblock In {\em {IEEE}/{CVF} {Conference} on {Computer} {Vision} and
  {Pattern} {Recognition} ({CVPR})}. IEEE, 2019.

\bibitem{zela_understanding_2020}
Arber Zela, Thomas Elsken, Tonmoy Saikia, Yassine Marrakchi, Thomas Brox, and
  Frank Hutter.
\newblock Understanding and {Robustifying} {Differentiable} {Architecture}
  {Search}.
\newblock In {\em International {Conference} on {Learning} {Representations}
  (ICLR)}, 2019.

\bibitem{chen_stabilizing_2021}
Xiangning Chen and Cho-Jui Hsieh.
\newblock Stabilizing {Differentiable} {Architecture} {Search} via
  {Perturbation}-based {Regularization}.
\newblock In {\em International Conference on Machine Learning (ICML)}, 2020.

\bibitem{zhang_idarts_2021}
Miao Zhang, Steven~W. Su, Shirui Pan, Xiaojun Chang, Ehsan~M Abbasnejad, and
  Reza Haffari.
\newblock {iDARTS}: {Differentiable} {Architecture} {Search} with {Stochastic}
  {Implicit} {Gradients}.
\newblock In {\em {International} {Conference} on {Machine} {Learning} (ICML)},
  2021.

\bibitem{ho_population_2019}
Daniel Ho, Eric Liang, Ion Stoica, Pieter Abbeel, and Xi~Chen.
\newblock Population {Based} {Augmentation}: {Efficient} {Learning} of
  {Augmentation} {Policy} {Schedules}.
\newblock In {\em International {Conference} on {Machine} {Learning} ({ICML})},
  2019.

\bibitem{lim_fast_2019}
Sungbin Lim, Ildoo Kim, Taesup Kim, Chiheon Kim, and Sungwoong Kim.
\newblock Fast {AutoAugment}.
\newblock In {\em Advances in {Neural} {Information} {Processing} {Systems}
  ({NeurIPS})}, 2019.

\bibitem{hataya_faster_2020}
Ryuichiro Hataya, Jan Zdenek, Kazuki Yoshizoe, and Hideki Nakayama.
\newblock Faster {AutoAugment}: {Learning} {Augmentation} {Strategies} {Using}
  {Backpropagation}.
\newblock In {\em Computer {Vision} – {ECCV} 2020}. Springer International
  Publishing, 2020.

\bibitem{li_dada_2020}
Yonggang Li, Guosheng Hu, Yongtao Wang, Timothy Hospedales, Neil~M. Robertson,
  and Yongxin Yang.
\newblock {DADA}: {Differentiable} {Automatic} {Data} {Augmentation}.
\newblock In {\em {ECCV}}, 2020.

\bibitem{rommel2021cadda}
C{\'e}dric Rommel, Thomas Moreau, Joseph Paillard, and Alexandre Gramfort.
\newblock {CADDA: Class-wise Automatic Differentiable Data Augmentation for EEG
  Signals}.
\newblock In {\em International {Conference} on {Learning} {Representations}
  (ICLR)}, 2022.

\bibitem{liu_darts_2019}
Hanxiao Liu, Karen Simonyan, and Yiming Yang.
\newblock {DARTS}: {Differentiable} {Architecture} {Search}.
\newblock In {\em International {Conference} on {Learning} {Representations}
  (ICLR)}, 2019.

\bibitem{cohenc2016group}
Taco Cohen and Max Welling.
\newblock Group equivariant convolutional networks.
\newblock In {\em International Conference on Machine Learning (ICML)}. PMLR,
  2016.

\bibitem{Zaheer2017deep}
Manzil Zaheer, Satwik Kottur, Siamak Ravanbakhsh, Barnabas Poczos, Russ~R
  Salakhutdinov, and Alexander~J Smola.
\newblock Deep sets.
\newblock In {\em Advances in Neural Information Processing Systems (NeurIPS)},
  2017.

\bibitem{van_der_wilk_learning_2018}
Mark van~der Wilk, Matthias Bauer, S.~T. John, and James Hensman.
\newblock Learning {Invariances} using the {Marginal} {Likelihood}.
\newblock In {\em Advances in {Neural} {Information} {Processing} {Systems}
  (NeurIPS)}, 2018.

\bibitem{benton_learning_2020}
Gregory Benton, Marc Finzi, Pavel Izmailov, and Andrew~Gordon Wilson.
\newblock Learning {Invariances} in {Neural} {Networks}.
\newblock In {\em Advances in Neural Information Processing Systems (NeurIPS)},
  2020.

\bibitem{chen_group-theoretic_2020}
Shuxiao Chen, Edgar Dobriban, and Jane~H. Lee.
\newblock A {Group}-{Theoretic} {Framework} for {Data} {Augmentation}.
\newblock In {\em Advances in {Neural} {Information} {Processing} {Systems}
  ({NeurIPS})}, 2020.

\bibitem{schulman_gradient_2015}
John Schulman, Nicolas Heess, Theophane Weber, and Pieter Abbeel.
\newblock Gradient {Estimation} {Using} {Stochastic} {Computation} {Graphs}.
\newblock In {\em Advances in {Neural} {Information} {Processing} {Systems}
  (NeurIPS)}, 2015.

\bibitem{bengio_estimating_2013}
Yoshua Bengio, Nicholas Léonard, and Aaron~C. Courville.
\newblock Estimating or {Propagating} {Gradients} {Through} {Stochastic}
  {Neurons} for {Conditional} {Computation}.
\newblock {\em CoRR}, 2013.

\bibitem{grathwohl_backpropagation_2018}
Will Grathwohl, Dami Choi, Yuhuai Wu, Geoff Roeder, and David Duvenaud.
\newblock Backpropagation through the {Void}: {Optimizing} control variates for
  black-box gradient estimation.
\newblock In {\em International {Conference} on {Learning} {Representations}
  (ICLR)}, 2018.

\bibitem{schwabedal_addressing_2019}
Justus T.~C. Schwabedal, John~C. Snyder, Ayse Cakmak, Shamim Nemati, and
  Gari~D. Clifford.
\newblock Addressing {Class} {Imbalance} in {Classification} {Problems} of
  {Noisy} {Signals} by using {Fourier} {Transform} {Surrogates}.
\newblock {\em arXiv:1806.08675}, 2019.

\bibitem{wang_data_2018}
Fang Wang, Sheng-hua Zhong, Jianfeng Peng, Jianmin Jiang, and Yan Liu.
\newblock Data {Augmentation} for {EEG}-{Based} {Emotion} {Recognition} with
  {Deep} {Convolutional} {Neural} {Networks}.
\newblock In {\em {MultiMedia} {Modeling}}, volume 10705, pages 82--93.
  Springer International Publishing, 2018.
\newblock Series Title: Lecture Notes in Computer Science.

\bibitem{bouchacourt2021grounding}
Diane Bouchacourt, Mark Ibrahim, and Ari~S. Morcos.
\newblock Grounding inductive biases in natural images: invariance stems from
  variations in data.
\newblock In {\em Advances in Neural Information Processing Systems (NeurIPS)},
  2021.

\bibitem{krizhevsky2009learning}
Alex Krizhevsky, Geoffrey Hinton, et~al.
\newblock Learning multiple layers of features from tiny images.
\newblock 2009.

\bibitem{he2016identity}
Kaiming He, Xiangyu Zhang, Shaoqing Ren, and Jian Sun.
\newblock Identity mappings in deep residual networks.
\newblock In {\em Computer Vision -- ECCV 2016}, 2016.

\bibitem{he2016deep}
Kaiming He, Xiangyu Zhang, Shaoqing Ren, and Jian Sun.
\newblock Deep residual learning for image recognition.
\newblock In {\em {IEEE}/{CVF} {Conference} on {Computer} {Vision} and
  {Pattern} {Recognition} ({CVPR})}, 2016.

\bibitem{cubuk_randaugment_2020}
Ekin~D. Cubuk, Barret Zoph, Jonathon Shlens, and Quoc~V. Le.
\newblock Randaugment: {Practical} automated data augmentation with a reduced
  search space.
\newblock In {\em {IEEE}/{CVF} {Conference} on {Computer} {Vision} and
  {Pattern} {Recognition} {Workshops} ({CVPRW})}, pages 3008--3017. IEEE, 2020.

\bibitem{Inoue2018DataAB}
Hiroshi Inoue.
\newblock Data augmentation by pairing samples for images classification.
\newblock In {\em {IEEE}/{CVF} {Conference} on {Computer} {Vision} and
  {Pattern} {Recognition} ({CVPR})}, 2018.

\bibitem{Howard2014SomeIO}
Andrew~G. Howard.
\newblock Some improvements on deep convolutional neural network based image
  classification.
\newblock {\em CoRR}, abs/1312.5402, 2014.

\bibitem{Iber2007}
C.~Iber, S.~Ancoli-Israel, A.~Chesson, and S.~F Quan.
\newblock {The AASM Manual for the Scoring of Sleep and Associated Events:
  Rules, Terminology and Technical Specification}, 2007.

\bibitem{o2014montreal}
Christian O'reilly, Nadia Gosselin, Julie Carrier, and Tore Nielsen.
\newblock Montreal archive of sleep studies: an open-access resource for
  instrument benchmarking and exploratory research.
\newblock {\em Journal of sleep research}, 23(6):628--635, 2014.

\bibitem{kingma_adam_2015}
Diederik~P. Kingma and Jimmy Ba.
\newblock Adam: {A} {Method} for {Stochastic} {Optimization}.
\newblock In {\em {International} {Conference} on {Learning} {Representations}
  {(ICLR)}}, 2015.

\bibitem{loshchilov2017decoupled}
Ilya Loshchilov and Frank Hutter.
\newblock Decoupled weight decay regularization.
\newblock In {\em International {Conference} on {Learning} {Representations}
  ({ICLR})}, 2019.

\bibitem{eriba2019kornia}
E.~Riba, D.~Mishkin, E.~Rublee D.~Ponsa, and G.~Bradski.
\newblock {Kornia: an Open Source Differentiable Computer Vision Library for
  PyTorch}.
\newblock In {\em Winter Conference on Applications of Computer Vision}, 2020.

\bibitem{GramfortEtAl2013a}
Alexandre Gramfort, Martin Luessi, Eric Larson, Denis~A. Engemann, Daniel
  Strohmeier, Christian Brodbeck, Roman Goj, Mainak Jas, Teon Brooks, Lauri
  Parkkonen, and Matti~S. H{\"a}m{\"a}l{\"a}inen.
\newblock {{MEG}} and {{EEG}} data analysis with {{MNE}}-{{Python}}.
\newblock {\em Frontiers in Neuroscience}, 7(267):1--13, 2013.

\bibitem{braindecode}
Robin~Tibor Schirrmeister, Jost~Tobias Springenberg, Lukas Dominique~Josef
  Fiederer, Martin Glasstetter, Katharina Eggensperger, Michael Tangermann,
  Frank Hutter, Wolfram Burgard, and Tonio Ball.
\newblock Deep learning with convolutional neural networks for {EEG} decoding
  and visualization.
\newblock {\em Human Brain Mapping}, 2017.

\bibitem{akiba_optuna_2019}
Takuya Akiba, Shotaro Sano, Toshihiko Yanase, Takeru Ohta, and Masanori Koyama.
\newblock Optuna: {A} {Next}-generation {Hyperparameter} {Optimization}
  {Framework}.
\newblock In {\em Proceedings of the 25th {ACM} {SIGKDD} {International}
  {Conference} on {Knowledge} {Discovery} and {Data} {Mining}}, 2019.

\end{thebibliography}
}

\clearpage
\appendix


\counterwithin{figure}{section}
\counterwithin{table}{section}
\renewcommand{\thefigure}{\thesection.\arabic{figure}}
\renewcommand{\thetable}{\thesection.\arabic{table}}
\section{Experimental settings}

In all experiments, a grid-search was used to find the value of the regularization parameter $\lambda$ from \eqref{eq:learning-objective}.
In most cases, the grid of values used was $0.05, 0.1, 0.5$ and $1.$.
For the sleep stage classification experiment of \autoref{sec:mass}, stronger values were explored: $1, 5, 10, 50$.
\CedricRebuttal{An analysis for the CIFAR10 experiment of \autoref{sec:cifar10} can be found in \autoref{app:more-cifar}.}

\subsection{Mario-Iggy experiment -- \autoref{sec:mario}}
\label{app:setting-mario}

\begin{figure}[h]
    \centering
    \includegraphics[width=0.6\linewidth]{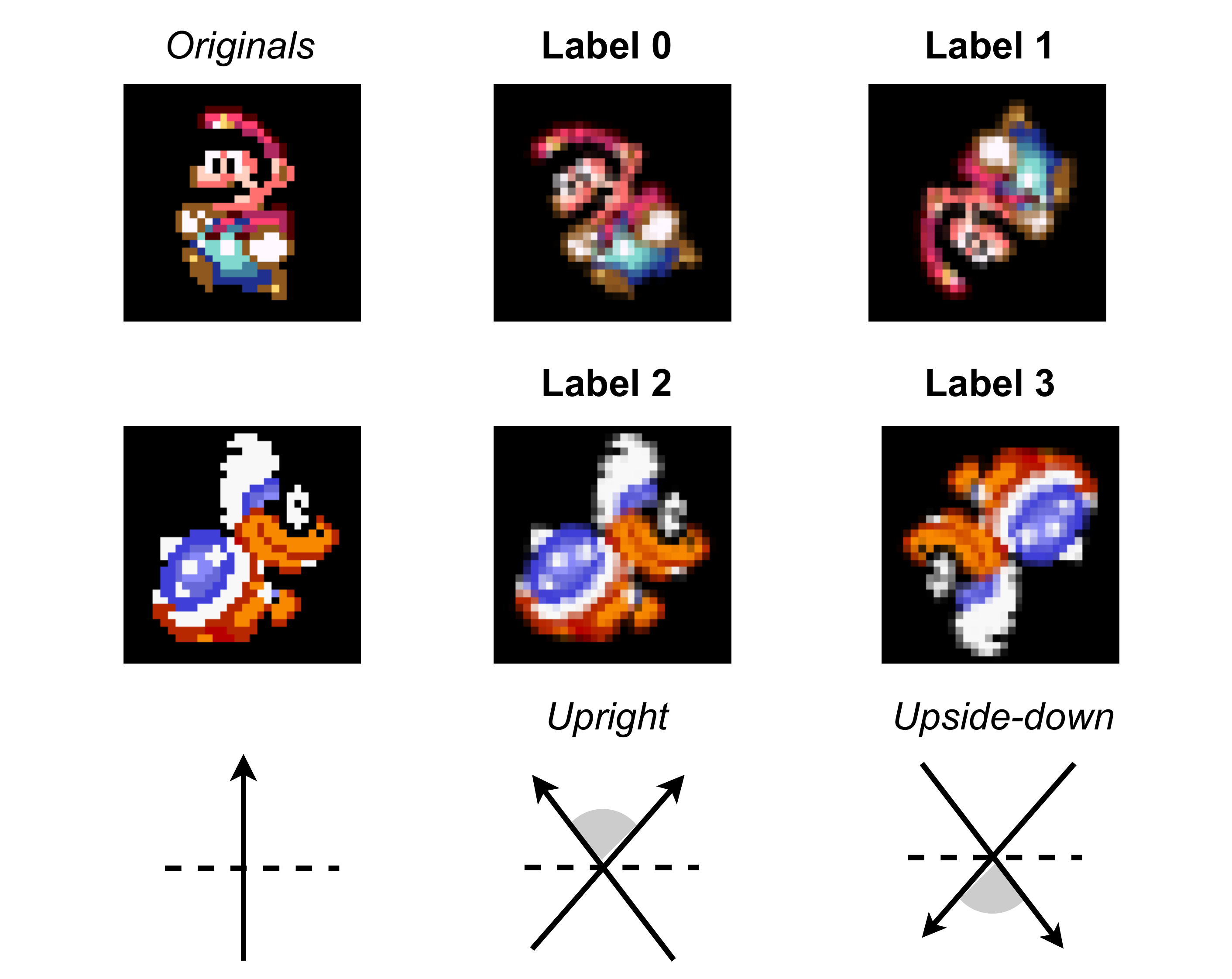}
    \caption{Illustration of the data generation process for the Mario-Iggy experiment.}
    \label{fig:mario-data}
\end{figure}

The data was generated as described in \autoref{fig:mario-data} and \autoref{sec:mario}.
We used 10000 training examples and 5000 test examples, and sampled batches of 128 images, as in \cite{benton_learning_2020}.
The trunk model used for both Augerino and Augnet are described in \autoref{tab:mario-cnn}.
The official code from \cite{benton_learning_2020} was used for this experiment.
As in the original experiments, all models were trained for 20 epochs using Adam~\cite{kingma_adam_2015}, with $\beta_1=0.9$ and $\beta_2=0.999$.
The learning rate for Augerino was set to $10^{-2}$ and weight decay was set to $0.$
The regularization parameter was set to $\lambda=0.05$ (medium according to \cite{benton_learning_2020}).
For AugNet, we used a learning rate of $5 \times 10^{-4}$ and a regularization parameter of $0.5$, together with weight decay of $1$.
Augmentations were all initialized with magnitudes equivalent to $\pi/8$ for AugNet and Augerino. Augmentations were also initialized with uniform weights for AugNet.

\begin{figure}[h]
    \centering
    \includegraphics[width=0.6\linewidth]{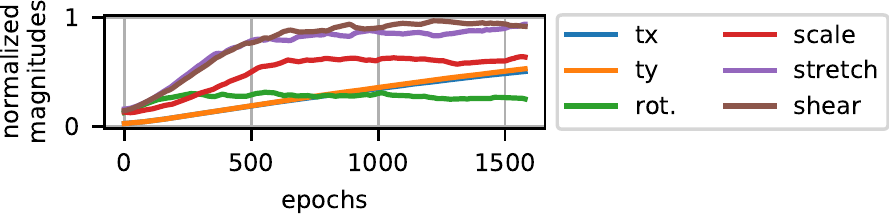}
    \caption{Magnitudes learned by Augerino are not sparse.}
    \label{fig:mario-augerino-mags}
\end{figure}

\begin{figure}[t]
    \centering
    \begin{subfigure}[b]{0.3\linewidth}
        \includegraphics[width=\linewidth]{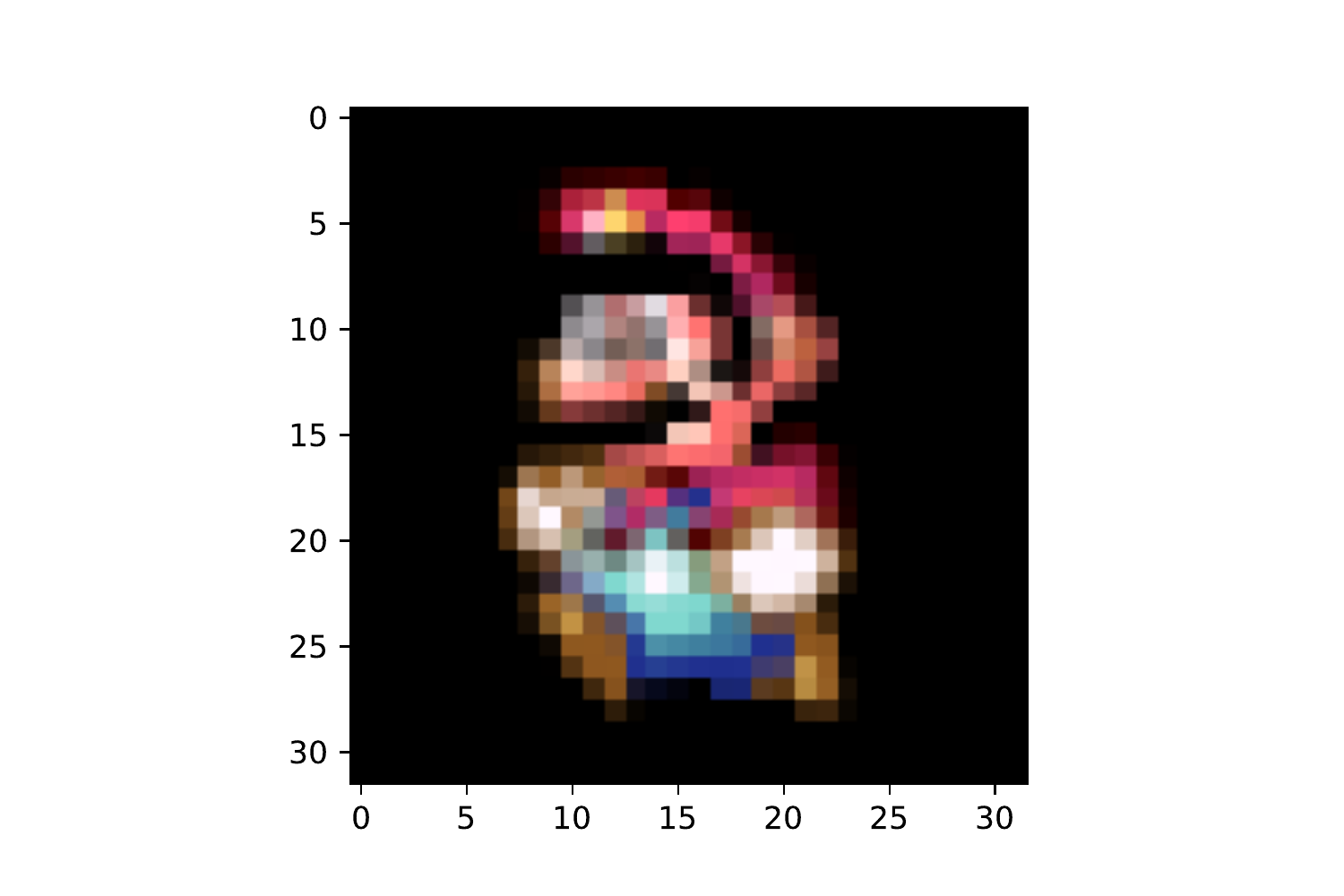}
        \caption{}
    \end{subfigure}
    \hfill
    \begin{subfigure}[b]{0.3\linewidth}
        \includegraphics[width=\linewidth]{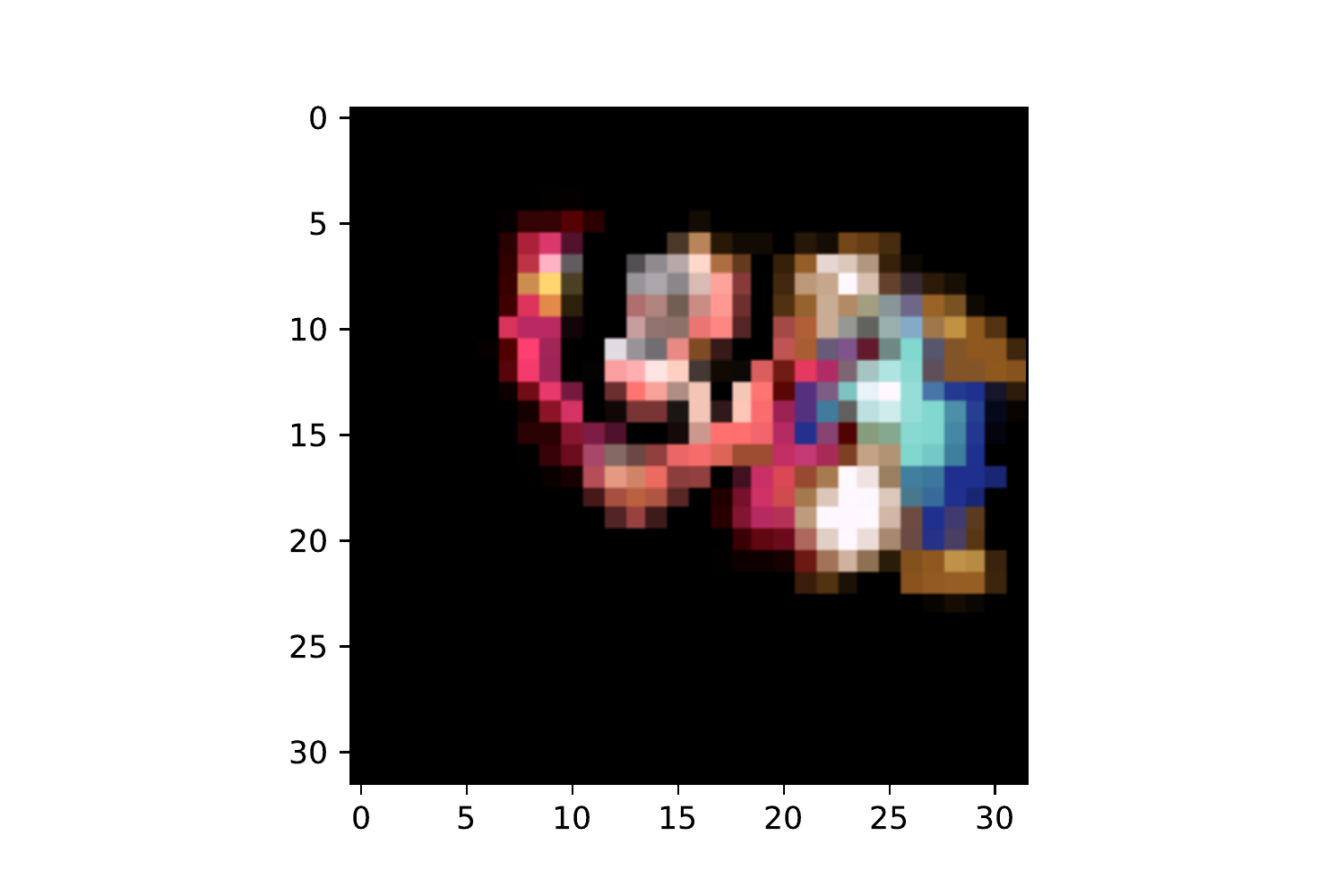}
        \caption{}
    \end{subfigure}
    \hfill
    \begin{subfigure}[b]{0.3\linewidth}
        \includegraphics[width=\linewidth]{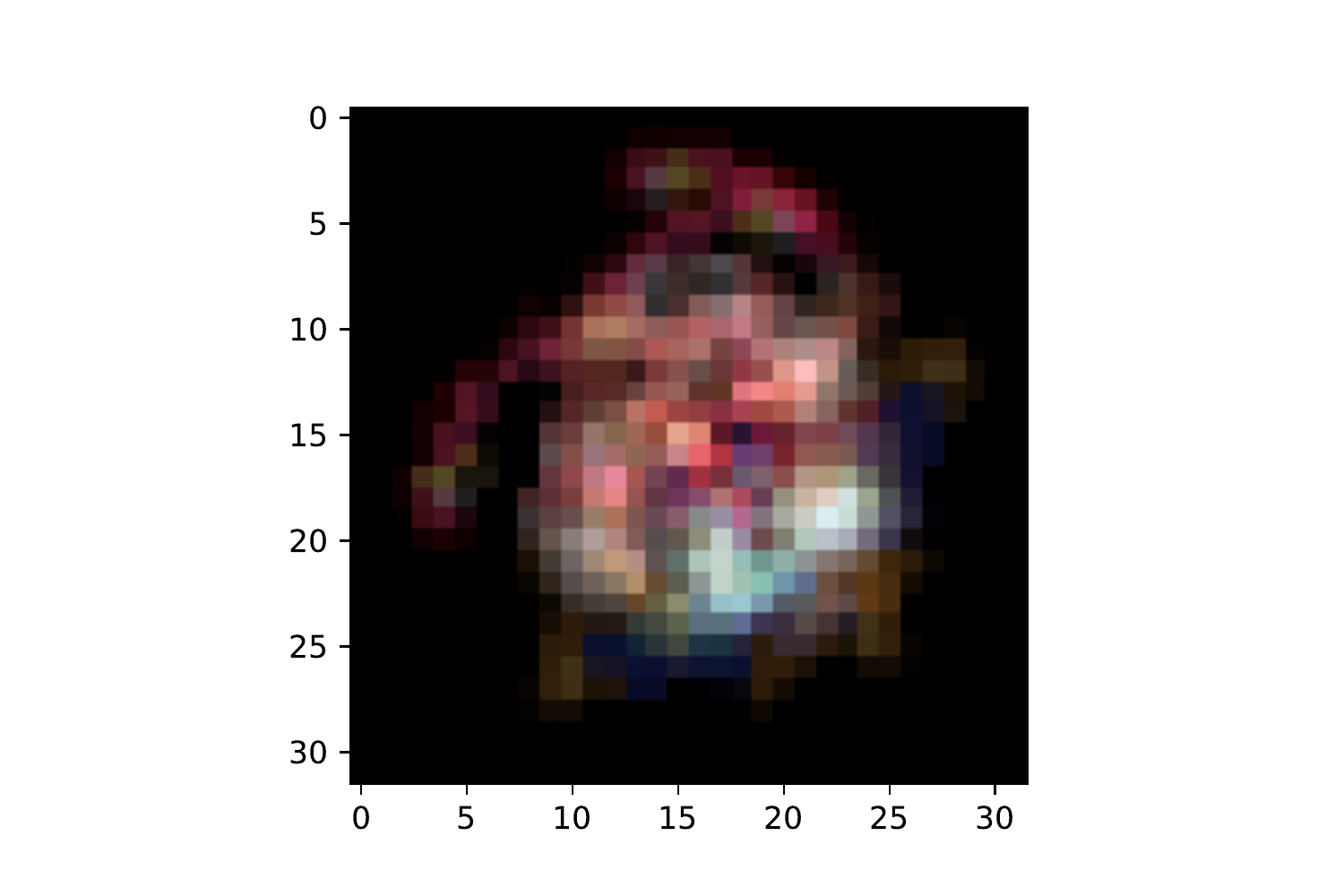}
        \caption{}
    \end{subfigure}
    \caption{\textbf{(a)} Original picture, \textbf{(b)} picture transformed with a sequence of \texttt{flip, rotate, crop}, \textbf{(c)} picture transformed with a convex sum of them.
    }
\end{figure}

\begin{table}[h]
    \centering
    \begin{tabular}{c|ccccc}
         &  \textbf{layer} & \textbf{\# filters} & \textbf{size} & \textbf{stride} & \textbf{batch norm} \\ \hline
        1 & Conv2D & 32 & 3 & 1 & yes\\ 
          & ReLU & & & & \\ \hline
        2 & Conv2D & 64 & 3 & 1 & yes\\
          & ReLU & & & & \\
          & MaxPool & & 2 & & \\ \hline
        3 & Conv2D & 128 & 3 & 1 & yes\\
          & ReLU & & & & \\
          & MaxPool & & 2 & & \\ \hline
        4 & Conv2D & 256 & 3 & 1 & yes\\
          & ReLU & & & & \\
          & MaxPool & & 2 & & \\ \hline
        5 & MaxPool & & 4 & 1 & \\
          & FC & & & &\\
          & Softmax & & & &
    \end{tabular}
    \medskip
    \caption{Convolutional neural network architecture used in experiments of \autoref{sec:mario}.}
    \label{tab:mario-cnn}
\end{table}

\subsection{Sinusoids experiment -- \autoref{sec:freq-pred}}
\label{app:setting-sinus}

\begin{figure*}[h]
    \centering
    \includegraphics[width=\linewidth]{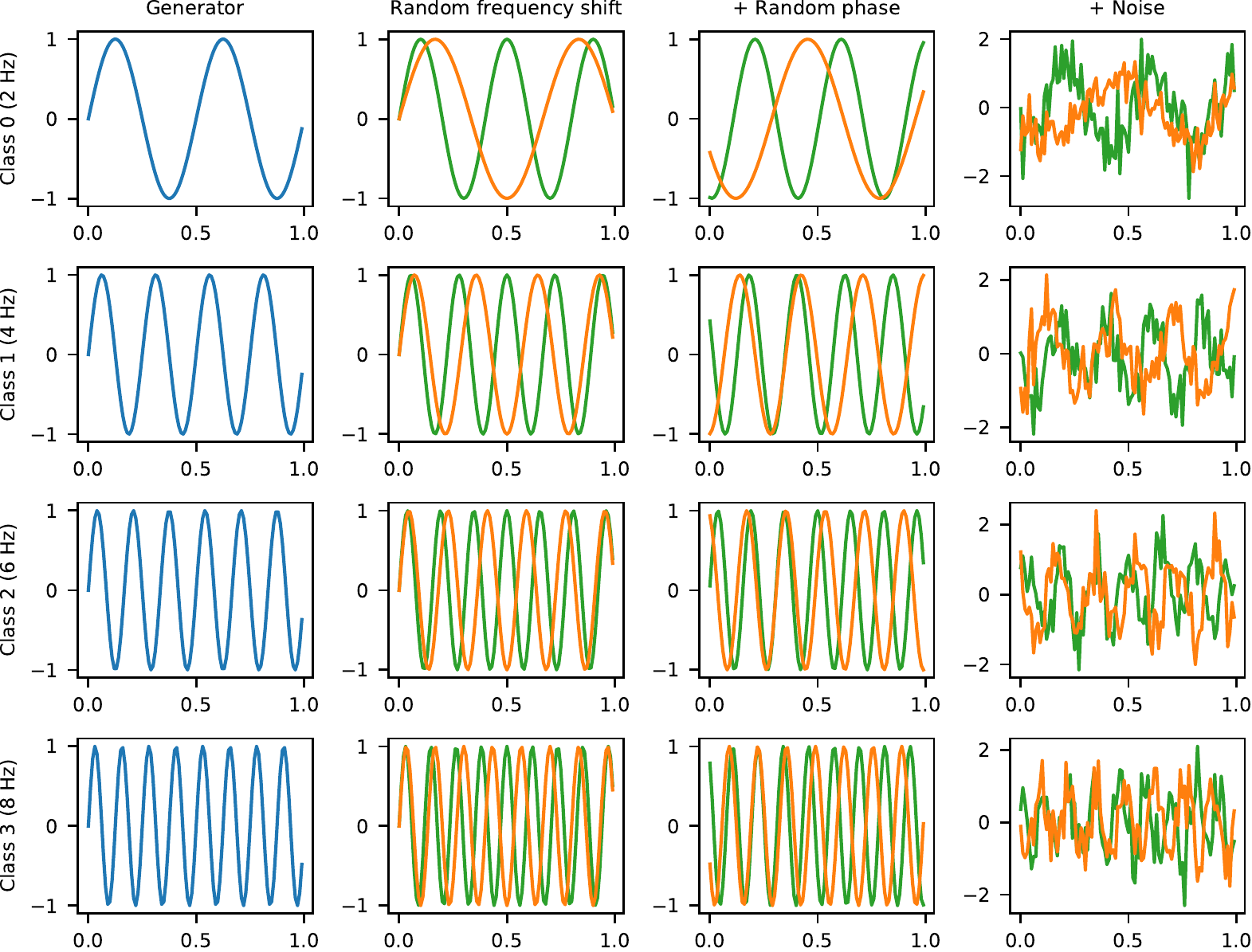}
    \caption{Illustration (cropped to 1 second) of waves composing each class of the simulated dataset used in the sinusoids and model capacity experiments.}
    \label{fig:sinusoids-dataset}
\end{figure*}

The data was generated as described in \autoref{fig:sinusoids-dataset} and \autoref{sec:freq-pred}.
We used 400 training examples and 200 test examples, and sampled batches of 32 waves.
The trunk model used for Augnet is described in \autoref{tab:sinus-cnn}.
All models were trained for 50 epochs using Adam~\cite{kingma_adam_2015}.
We used a learning rate of $10^{-2}$ and a regularization parameter of $0.2$, together with weight decay of $10^{-4}$.
Augmentations were all initialized with magnitudes $\mu=0$ and uniform weights.
For the experiment of \autoref{sec:model-capacity}, we set the number of copies to $C=10$ at inference, while it was set to $C=4$ for all other experiments.
Moreover, $\lambda$ was set to $0.8$ and initial magnitudes set to $0.05$ in \autoref{sec:model-capacity}.

\begin{table}[h]
    \centering
    \begin{tabular}{c|ccccc}
         &  \textbf{layer} & \textbf{\# filters} & \textbf{size} & \textbf{stride} & \textbf{batch norm} \\ \hline
        1 & Conv2D & 2 & 3 & 1 & yes\\
          & ReLU & & & & \\ \hline
        2 & Conv2D & 2 & 3 & 1 & yes\\
          & ReLU & & & & \\
          & MaxPool & & 2 & & \\ \hline
        3 & GlobalPool & & & & \\
          & FC & & & &\\
          & Softmax & & & &
    \end{tabular}
    \medskip
    \caption{Convolutional neural network architecture used in experiments of \autoref{sec:freq-pred}}
    \label{tab:sinus-cnn}
\end{table}

\subsection{CIFAR10 experiment -- \autoref{sec:cifar10}}
\label{app:setting-cifar}

\Cedric{The official code from \cite{benton_learning_2020} was used for this experiment.
It is worth noting that Augerino's official code used a smaller trunk model with 13 layers, which led to poor performances. We hence replaced it by a Pre-activate ResNet18, which significantly improved performances.
The official code also did not implement any cosine annealing despite what was reported in \cite{benton_learning_2020}.
We were hence not able to reproduce their results exactly despite our efforts.}

Following \cite{benton_learning_2020}, we used batches of 128 images.
All models were trained for \Cedric{300}
epochs using Adam~\cite{kingma_adam_2015}
\Cedric{with decoupled weight-decay~\cite{loshchilov2017decoupled}.}
Also, as described in \cite{benton_learning_2020}, we used a cosine annealing scheduler with
\Cedric{period $T=300$ epochs.}
\Cedric{For AugNet, Augerino and the baseline with fixed augmentations, the following normalization was used as a preprocessing on augmented data for the whole dataset:}
centering by $(0.485, 0.456, 0.406)$ and scaling by $(0.229, 0.224, 0.225)$.
\Cedric{All models were trained with a learning rate set to $10^{-3}$. Weight-decay was globally set to $2\times10^{-2}$ for the trunk model and $0$ for the augmentation modules of both Augerino and AugNet.}
The regularization parameter was set to 
\Cedric{$\lambda=0.05$}
for Augerino as in \cite{benton_learning_2020},
\Cedric{while AugNet's regularization was set to $0.5$ for the first 40 epochs and scaled down to $0.05$ for the rest of the training.}
Augmentations were all initialized with magnitudes set to $0.5$ for both Augerino and AugNet, with uniform weights for the latter.
In this experiments, differentiable augmentations were implemented using the \textsc{Kornia} package~\cite{eriba2019kornia}, as well as the official code of Faster AutoAugment~\cite{hataya_faster_2020}. All trainings were carried on single Tesla V100 GPUs.

\subsection{MASS experiment -- \autoref{sec:mass}}
\label{app:setting-mass}

\paragraph{Dataset}
The public dataset MASS - Session 3 \cite{o2014montreal} was used for this purpose (more details in \autoref{app:setting-mass}).
It corresponds to 62 nights, each one coming from a different subject.
Out of the 20 available EEG channels, referenced with respect to the A2 electrode, we used 6 (C3, C4, F3, F4, O1, O2).
As done by \cite{rommel2021cadda}, both the training and validation sets consisted of 24 nights each, and the test set contained 12 nights.

\paragraph{Architecture}
For all EEG experiments, learning was carried using the convolutional network proposed in~\cite{chambon_deep_2017}, whose architecture is described on \autoref{tab:cnn_architecture}.
The initial number of channels $C$ was set to 8.
The first layers (1-4) implements a spatial filter, computing virtual channels through a linear combination of the original input channels.
Then, layers 5 to 9 correspond to a standard convolutional feature extractor and last layers implement a simple classifier.
\begin{table*}[ht!]
\centering
\def\arraystretch{1.2}
\scriptsize
\begin{tabular}{l|lllllll}
& \textbf{layer} & \textbf{\# filters} & \textbf{\# params}    & \textbf{size} & \textbf{stride} & \textbf{output dim.} & \textbf{activation} \\ \hline 
1              & Input               &                     &                       &               &                 & (C, T)                    &                     \\
2              & Reshape             &                     &                       &               &                 & (C, T, 1)                 &                     \\
3              & Conv2D      & C                   & C * C                 & (C, 1)        & (1, 1)          & (1, T, C)                 & Linear              \\
4              & Permute             &                     &                       &               &                 & (C, T, 1)                 &                     \\
5              & Conv2D      & 8                   & 8 * 64 + 8            & (1, 64)       & (1, 1)          & (C, T, 8)                 & Relu                \\
6              & Maxpool2D       &                     &                       & (1, 16)       & (1, 16)         & (C, T // 16, 8)           &                     \\
7              & Conv2D      & 8                   & 8 * 8 * 64 + 8        & (1, 64)       & (1, 1)          & (C, T // 16, 8)           & Relu                \\
8              & Maxpool2D       &                     &                       & (1, 16)       & (1, 16)         & (C, T // 256, 8)           &                    \\
9              & Flatten             &                     &                       &               &                 & (C * (T // 256) * 8)       &                    \\
10             & Dropout (50\%)      &                     &                       &               &                 & (C * (T // 256) * 8)       &                    \\
11             & Dense               &                     & 5 * (C * T // 256 * 8) &               &                 & 5                         & Softmax
\end{tabular}
\vspace{3mm}
\caption{Detailed architecture from \cite{chambon_deep_2017}, where $C$ is the number of EEG channels and $T$ the time series length.} \label{tab:cnn_architecture}
\end{table*}
\paragraph{Training hyperparameters}
The optimizer used for all models was Adam with a learning rate of $10^{-3}$, $\beta_1=0.$ and $\beta_2=0.999$.
At most 300 epochs were used for training, with a batch size of 16.
Early stopping was implemented with a patience of 30 epochs.
For ADDA, the policy learning rate was set to $5 \times 10^4$ based on a grid-search carried using the validation set.
For AugNet, the regularization parameter was set to $\lambda=10$.
Balanced accuracy was used as performance metric using the inverse of original class frequencies as balancing weights. The \textsc{MNE-Python}~\cite{GramfortEtAl2013a} and \textsc{Braindecode} software~\cite{braindecode} were used to preprocess and learn on the EEG data.
Training was carried on single Tesla V100 GPUs.

\paragraph{Augmentations considered}
The $13$ operations considered are listed in \autoref{tab:mass-augmentations}. A detailed explanation of their implementation can be found in the appendix of \cite{rommel2021cadda}. While all this augmentations were used by gradient-free algorithms, \texttt{bandstop filter} was not included in the differentiable strategies (Faster AA, DADA, ADDA, AugNet) because we did not implement a differentiable relaxation of it.
All augmentations used came from the \textsc{Braindecode} package~\cite{braindecode}.
AutoAugment was implemented replacing the PPO searcher with a TPE searcher (as in Fast AutoAugment). The \textsc{Optuna} package was used for that matter \cite{akiba_optuna_2019}.

\begin{table}[h]
    \centering
    \begin{tabular}{c|c|c}
        \textbf{type} & \textbf{transformation} & \textbf{range} \\ \hline
        Time & \texttt{time reverse} & \\
         & \texttt{time masking} & 0-200 samples \\
         & \texttt{Gaussian noise} & 0-0.2 std \\ \hline
        Frequency & \texttt{FT-surrogate} & 0-$2\pi$ \\
         & \texttt{frequency shift} & 0-5 Hz \\
         & \texttt{bandstop filter} & 0-2 Hz \\ \hline
        Sensors & \texttt{sign flip} & \\
         & \texttt{channels symmetry} & \\
         & \texttt{channels shuffle} & 0-1 \\
         & \texttt{channels dropout} & 0-1 \\
         & \texttt{rotations x-y-z} & 0-30 degrees \\
    \end{tabular}
    \medskip
    \caption{Augmentations considered in experiment \ref{sec:mass}. The range column corresponds to the values to which are mapped the magnitudes $\mu=0$ and $\mu=1$.}
    \label{tab:mass-augmentations}
\end{table}

\section{Complementary results}
\subsection{Sinusoids experiments -- \autoref{sec:freq-pred} and \ref{sec:model-capacity}}
\label{app:more-sinusoids}

Hereafter we provide further results and extensions to the experiment of \Autoref{sec:freq-pred, sec:model-capacity}.

\begin{figure}[h]
    \begin{subfigure}[b]{0.49\linewidth}
        \centering
        \includegraphics[width=\linewidth]{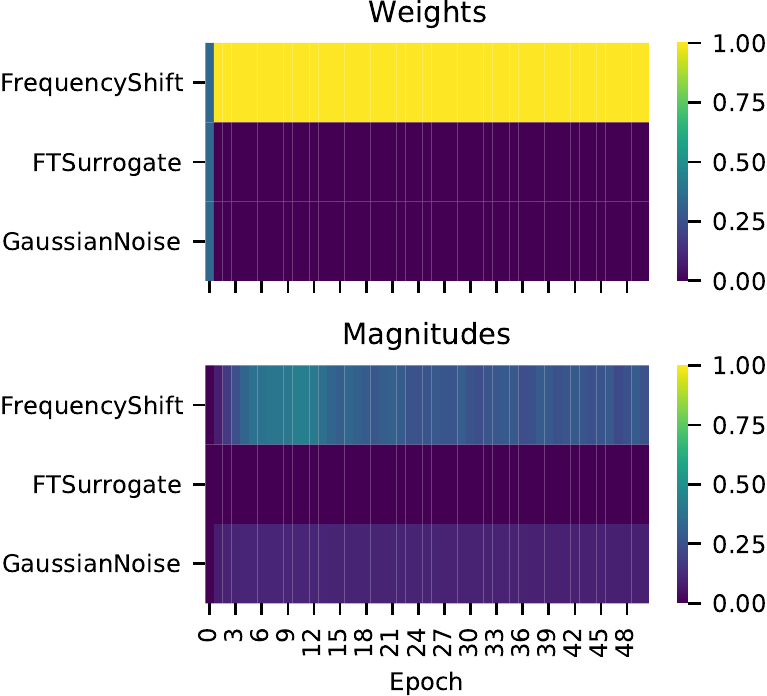}
        \caption{}
        \label{fig:sin-params-heatmap}
    \end{subfigure}
    \hfill
    \begin{subfigure}[b]{0.49\linewidth}
        \centering
        \includegraphics[width=\linewidth]{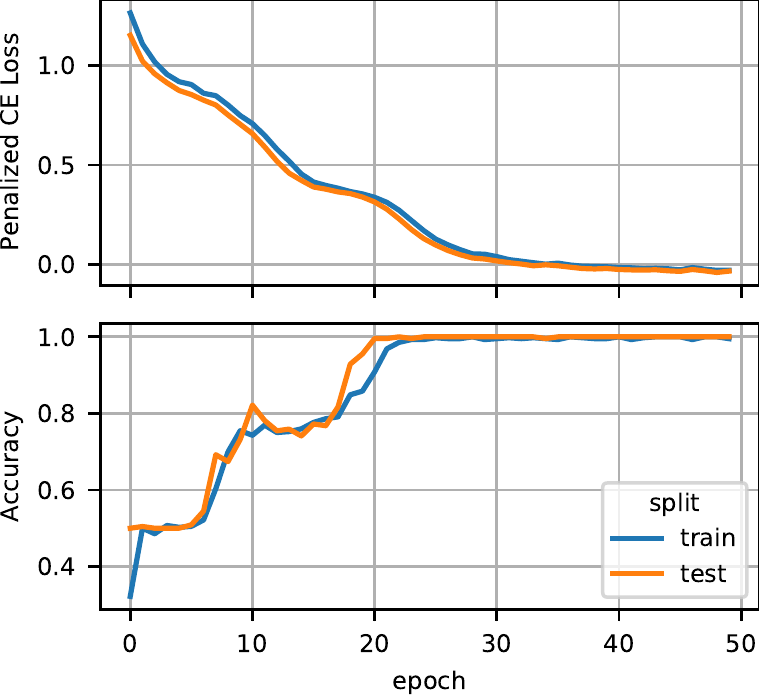}
        \caption{}
        \label{fig:sin-training}
    \end{subfigure}
    \caption{
    \textbf{(a)} Evolution of learned weights and magnitudes in the sinusoids experiment \autoref{sec:freq-pred}. AugNet quickly learns to maximize the frequency shift invariance and drop the others.
    \textbf{(b)} Loss and accuracy during AugNet training in the sinusoids experiment \autoref{sec:freq-pred}.
    }
\end{figure}

\Cedric{One may find the evolution of the augmentation module weights $w$ during training in the top-frame of \autoref{fig:sin-params-heatmap}.
We see that the correct \texttt{FrequencyShift} invariance is selected after only one epoch, its magnitude being correctly tuned as already shown on \autoref{fig:learned-shift}.}

\begin{figure*}[h]
    \centering
    \includegraphics[width=\linewidth]{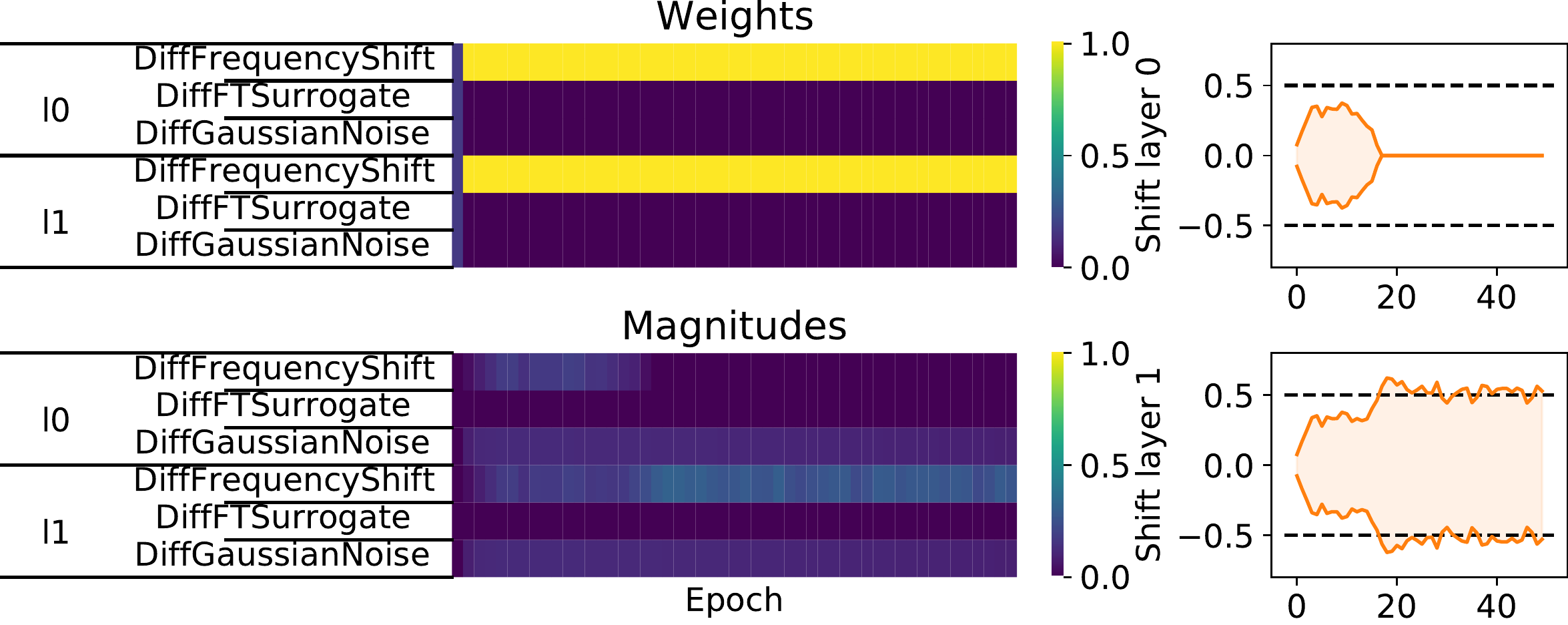}
    \caption{Weights and magnitudes learned by a model with 2 augmentation layers.}
    \label{fig:sin-params-heatmap-2l}
\end{figure*}

\begin{figure*}[h]
    \centering
    \includegraphics[width=\linewidth]{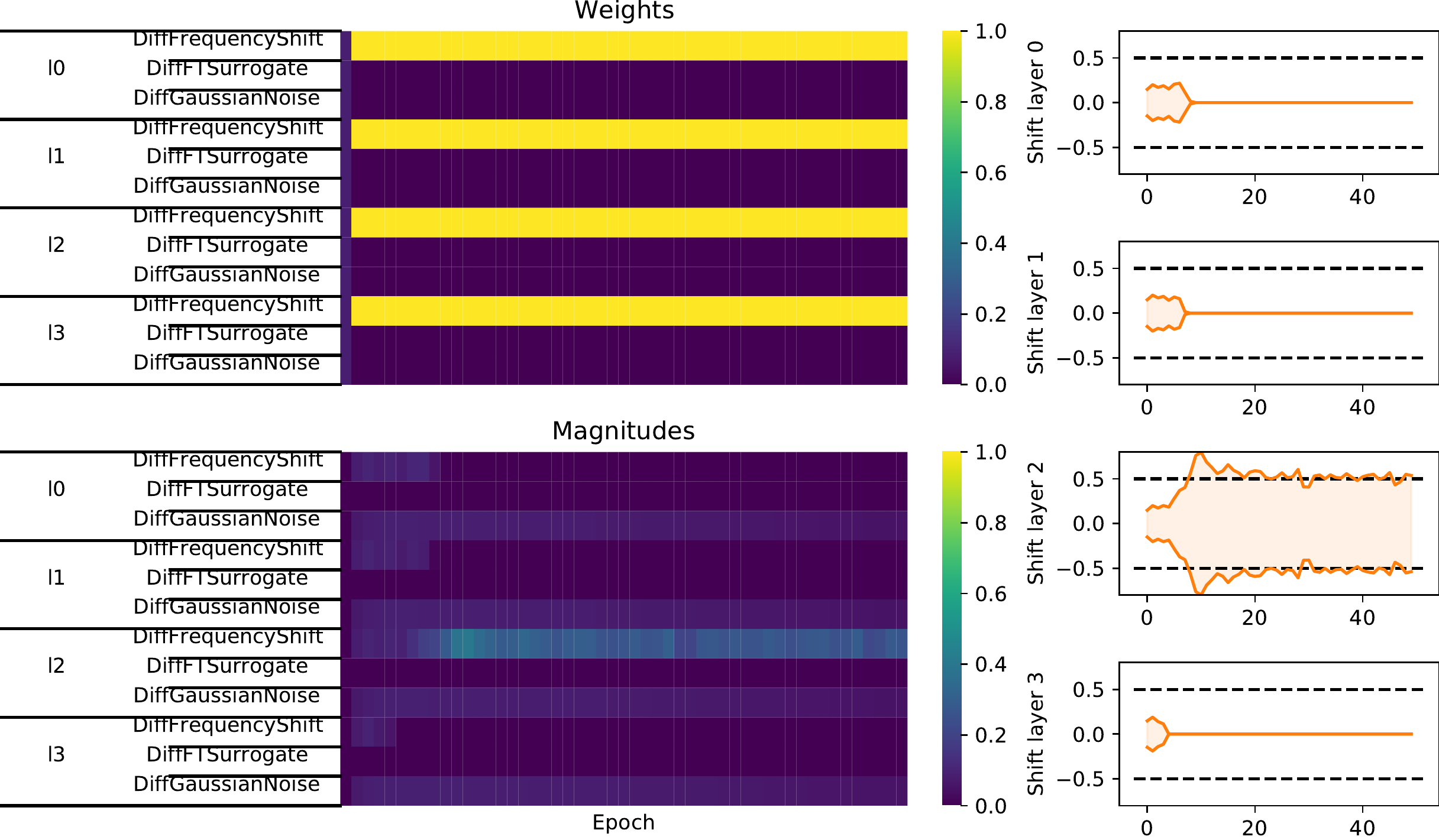}
    \caption{Weights and magnitudes learned by a model with 4 augmentation layers.}
    \label{fig:sin-params-heatmap-4l}
\end{figure*}

\paragraph{Sensitivity to number of layers}
Since the number of augmentation layers is a hyperparameter, we may wonder whether Augnet is robust to its value. Namely, we want to know what would happen if we trained an Augnet model with more augmentation layers  than there are invariances in the data.
Hence, we repeated the experiment from \autoref{sec:freq-pred} with 2 and 4 layers and plotted results on \Autoref{fig:sin-params-heatmap-2l, fig:sin-params-heatmap-4l} respectively.
The same training data and hyperparameters were used (\lcf \autoref{app:setting-sinus}).
We see in both cases that all layers select the correct transformation and only one gets a positive magnitude $\mu_i>0$, all other layers converging to the identity.
These results suggest that Augnet is indeed robust to the number of layers chosen and can correctly learn underlying invariances even when we decide to use ``too many'' of them.

\begin{figure*}[h]
    \centering
    \includegraphics[width=0.5\linewidth]{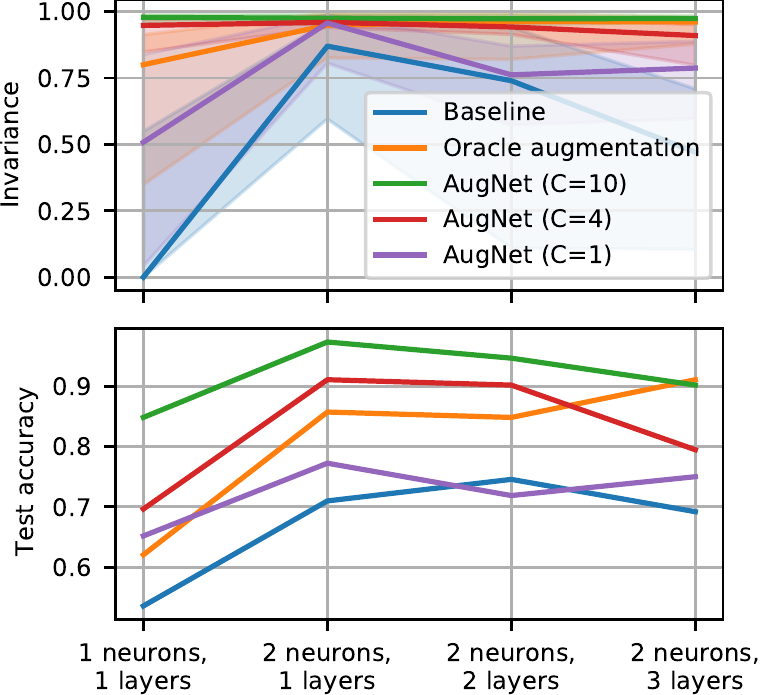}
    \caption{\CedricRebuttal{\textbf{Top:} Model invariance \eqref{eq:inv} to the true frequency shift in the data across different architectures. We report the median values across the test set, with a 75\% confidence interval. The greater $C$, the more invariant is AugNet.
    \textbf{Bottom:}
    By controlling the invariance, $C$ also controls AugNet's accuracy. AugNet has a performance close to the baseline with no augmentation if we don't average the model's predictions ($C=1$).}}
    \label{fig:capacity-study-c}
\end{figure*}

\CedricRebuttal{\paragraph{Sensitivity to $C$}
Another important hyperparameter to study in the context of this experiment is the number of copies $C$. Indeed, this parameter intuitively defines how well AugNet approximates the expectation in proposition \ref{eq:orbit-average}, and hence how invariant it is to the learned transformations. \autoref{fig:capacity-study-c} is an extended version of \autoref{fig:capacity-study}, where we have tested three different values $C$ at inference: 1, 4 and 10.
It confirms the intuition that the greater $C$, the more invariant is AugNet, and demonstrates that this has hence an important impact on performance.
The case $C=1$ works as a ``sanity check'' showing that we cannot do much better than the baseline with no augmentation if we don't average the model's predictions.}

\subsection{CIFAR10 experiment -- \autoref{sec:cifar10}}
\label{app:more-cifar}

\begin{figure}[h]
\begin{subfigure}[b]{0.49\linewidth}
        \centering
        \includegraphics[width=\linewidth]{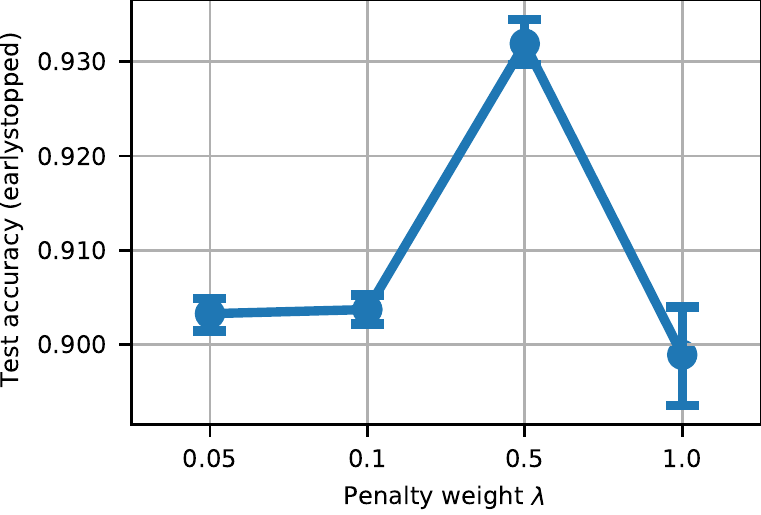}
        \caption{}
        \label{fig:lambda-vs-perf}
    \end{subfigure}
    \hfill
    \begin{subfigure}[b]{0.49\linewidth}
        \centering
        \includegraphics[width=\linewidth]{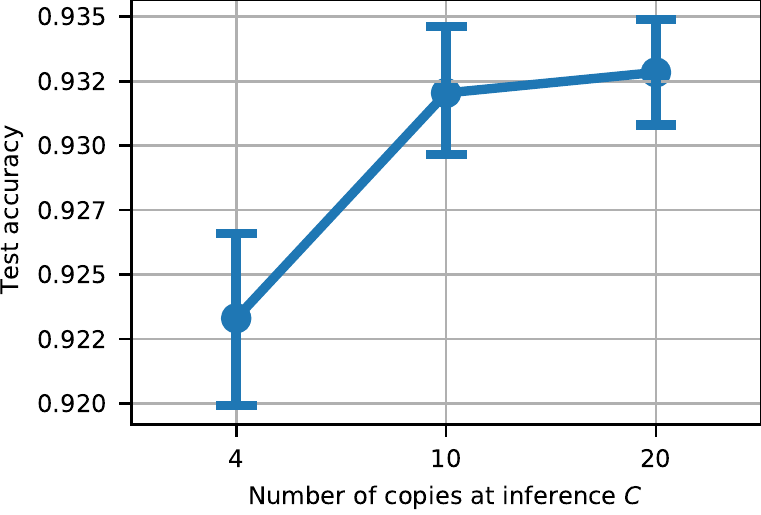}
        \caption{}
        \label{fig:c-vs-perf}
    \end{subfigure}
    \caption{\CedricRebuttal{ \textbf{(a)} Test accuracy on CIFAR10 for different penalty weights $\lambda$. $C$ is set to 20 in this experiment and all other details are given in \autoref{app:setting-cifar}.
    \textbf{(b)} Test accuracy on CIFAR10 for different number of copies $C$ at inference. All points correspond to the same model trained with $C=1$ at training and evaluated with different values of $C$ at inference.}
    }
    \label{fig:lambda-c-perf}
\end{figure}

\begin{figure}[h]
    \centering
    \includegraphics[width=0.5\linewidth]{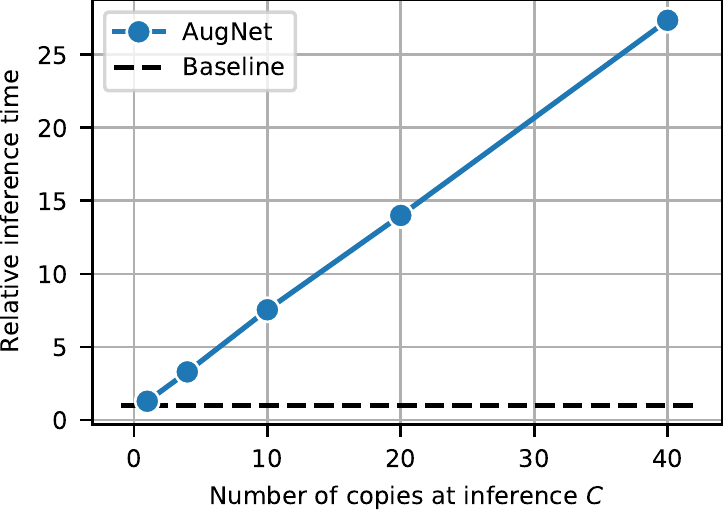}
    \caption{\CedricRebuttal{
    Average forward-pass time for different values of $C$ at inference.
    Error shades of 75\% confidence over batches are depicted, but very small.
    Results are normalized by the forward-pass time of the trunk model alone $f$, represented by the black dashed line.
    }
    \label{fig:c-vs-inf-time}
    }
\end{figure}

\CedricRebuttal{
\paragraph{Sensitivity analysis to $C$ and $\lambda$}
AugNet introduces two new hyperparameters: the penalty weight $\lambda$ and the number of copies $C$.
\autoref{fig:lambda-c-perf} shows a sensitivity analysis of the final performance on CIFAR10 (\autoref{sec:cifar10}) for both these hyperparameters.
We see on \autoref{fig:c-vs-perf} here again that larger values of $C$ yield better performances, as seen for the sinusoids simulated experiment in \autoref{fig:capacity-study-c}.
However, increasing the number of copies $C$ at inference also comes with a computation time that increases linearly, as shown on \autoref{fig:c-vs-inf-time}.
}

\begin{figure}[h]
    \centering
    \includegraphics[width=\linewidth]{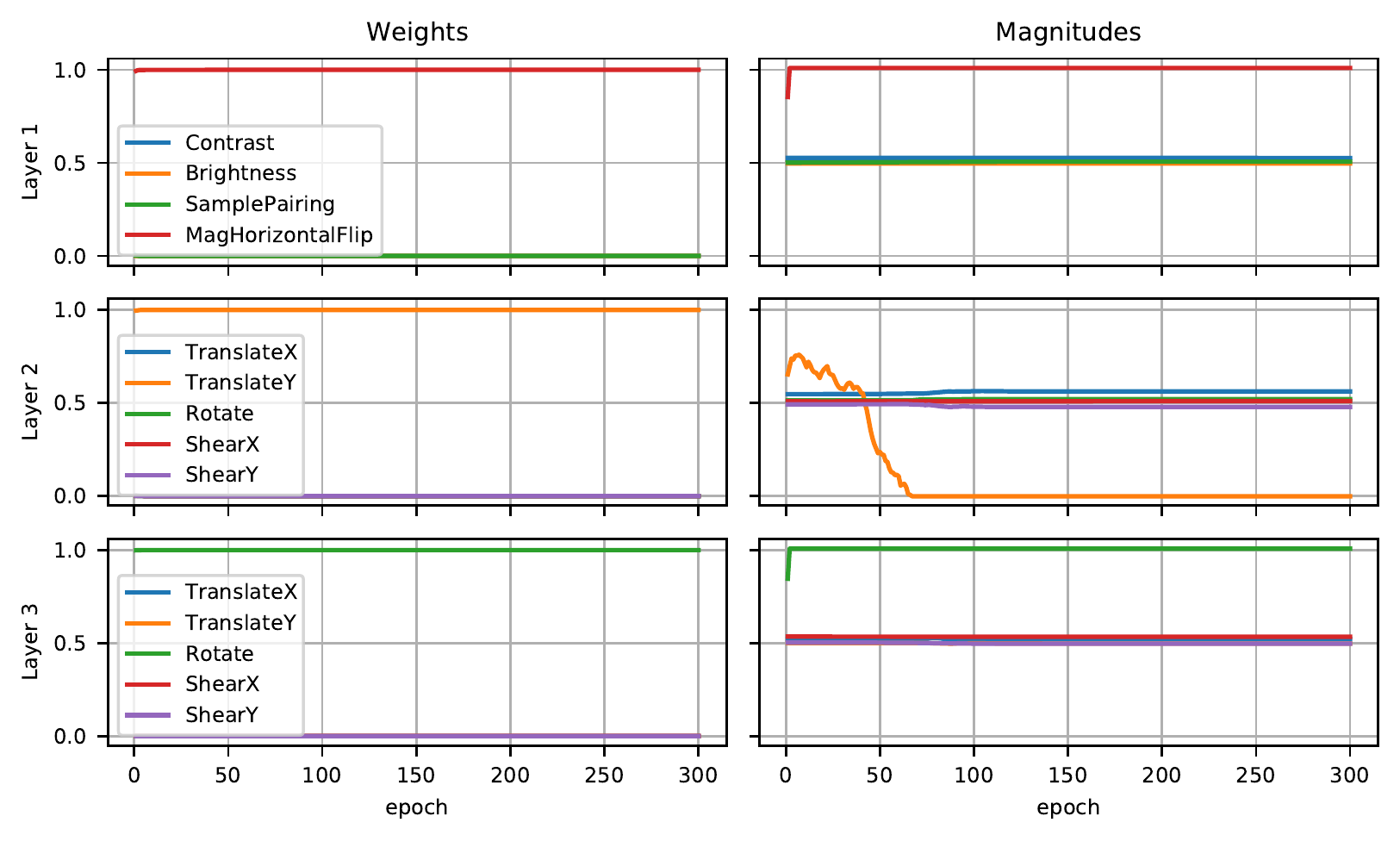}
    \caption{Learned invariances for one run of the CIFAR10 experiment. Very similar results were obtained for another run, an both led to the best performances.}
    \label{fig:cifar-selection-seed29}
\end{figure}

\begin{figure}[h]
    \centering
    \includegraphics[width=\linewidth]{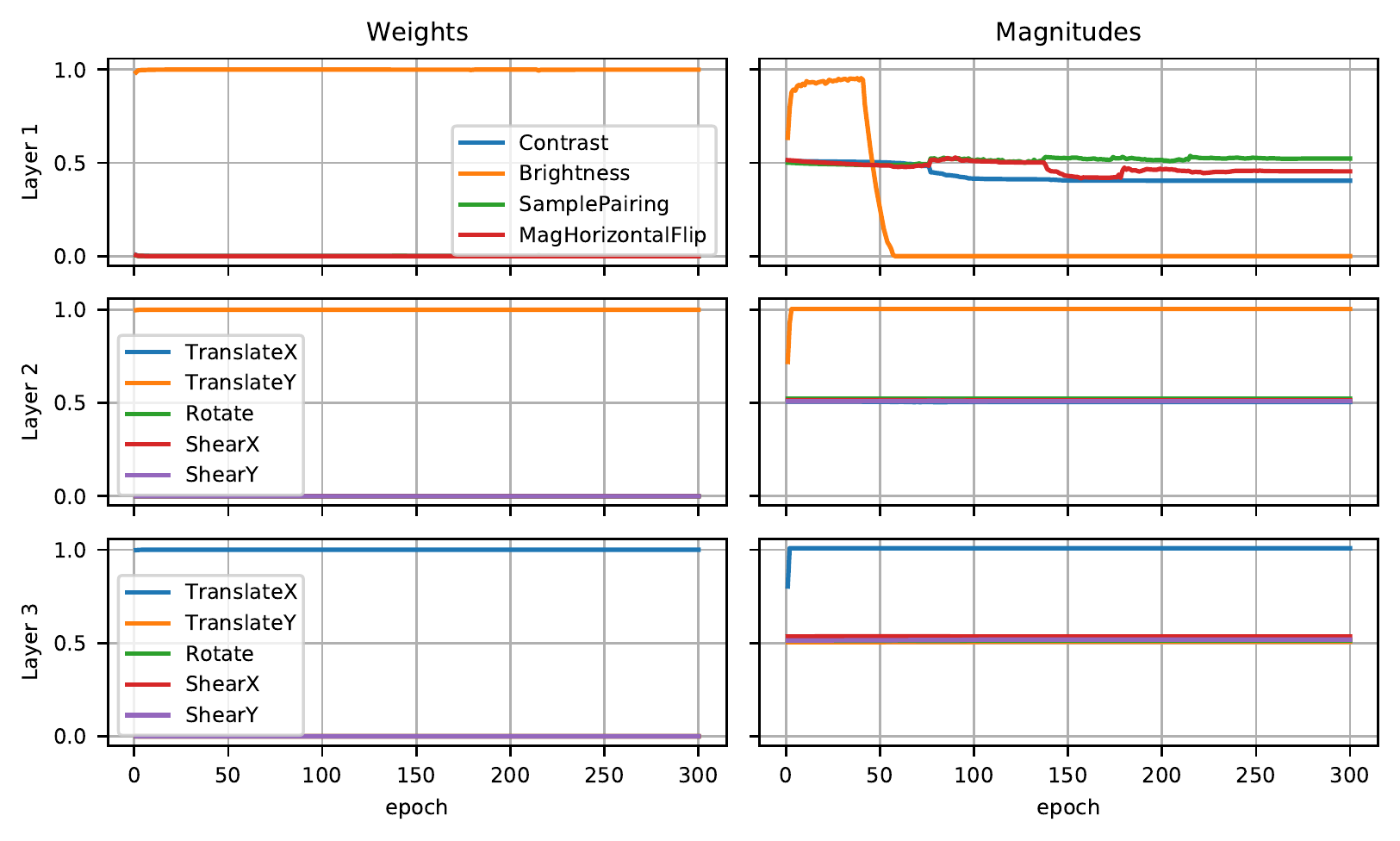}
    \caption{Learned invariances for another run of the CIFAR10 experiment. Very similar results were obtained for three other runs.}
    \label{fig:cifar-selection-seed1}
\end{figure}

\paragraph{Learned augmentations and model invariance}
\Autoref{fig:cifar-selection-seed29, fig:cifar-selection-seed1} show which invariances were learned for two of the five runs of the CIFAR10 experiment from \autoref{sec:cifar10}.
We see on the first case that layer 1 learned the \texttt{HorizontalFlip}, which is one of the two augmentations used in the experiment baseline leading to the state-of-the-art results.
We indeed got the best test accuracies
\CedricRebuttal{(93.4\% and 94.0\%)}
for the two runs which led to the selection depicted on \autoref{fig:cifar-selection-seed29} compared to
\CedricRebuttal{92.7 - 93.0\%}
for the other three runs with selections similar to \autoref{fig:cifar-selection-seed1}.
We can also see that all runs selected the \texttt{translate-y} invariance (and three of them selected \texttt{translate-x}), which is equivalent to the \texttt{random-crop} also used by the baseline with fixed augmentation.

\begin{figure}[h]
    \begin{subfigure}[b]{0.49\linewidth}
        \centering
        \includegraphics[width=\linewidth]{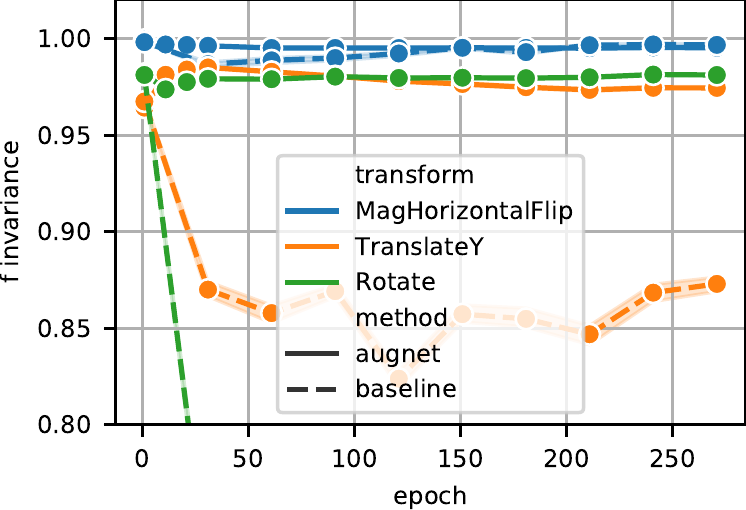}
        \caption{}
        \label{fig:cifar-inv-seed29}
    \end{subfigure}
    \hfill
    \begin{subfigure}[b]{0.49\linewidth}
        \centering
        \includegraphics[width=\linewidth]{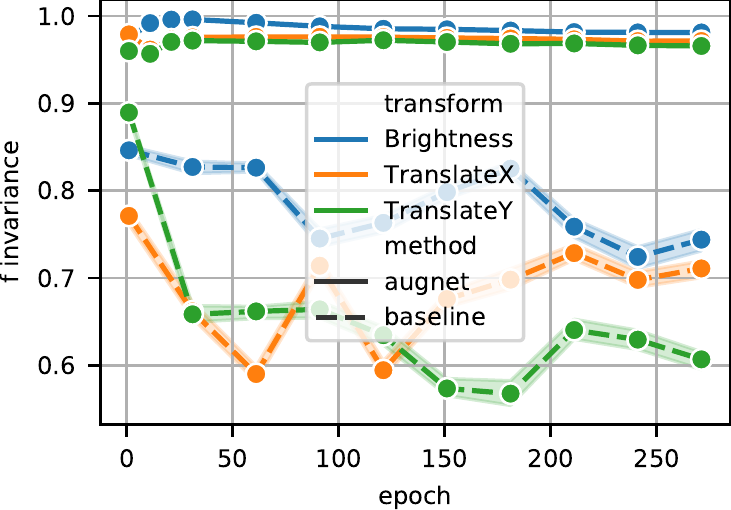}
        \caption{}
        \label{fig:cifar-inv-seed1}
    \end{subfigure}
    \caption{\CedricRebuttal{ 
    \textbf{(a)} Invariance \eqref{eq:inv} of AugNet and fixed augmentation baseline to augmentations selected by AugNet during the run from \autoref{fig:cifar-selection-seed29}.
    AugNet remains invariant to \texttt{translate-y} even after its magnitude drops to 0.
    \textbf{(b)} Invariance \eqref{eq:inv} of AugNet and fixed augmentation baseline to augmentations selected by AugNet during the run from \autoref{fig:cifar-selection-seed1}.
    AugNet remains invariant to \texttt{brightness} even after its magnitude drops to 0.
    }
    }
    \label{fig:cifar-inv}
\end{figure}

\CedricRebuttal{
Regarding the fact that for some layers, the magnitude of the selected transformation drops to 0 at a certain point in training, note that this only means that the augmentation is not required anymore to ensure the necessary level of invariance.
Indeed, it is known since Population-based augmentations~\cite{ho_population_2019} and RandAugment~\cite{cubuk_randaugment_2020} that the best augmentation depends on the stage of training.
As shown on \autoref{fig:cifar-inv}, AugNet remains invariant to these augmentations after those moments, which means that the weights of the trunk model $f$ have learned the invariance and that there is hence no more need to sample it.
}

\subsection{MASS experiment -- \autoref{sec:mass}}
\label{app:more-mass}

\begin{figure}[t]
    \begin{subfigure}[b]{0.49\linewidth}
        \centering
        \includegraphics[width=\linewidth]{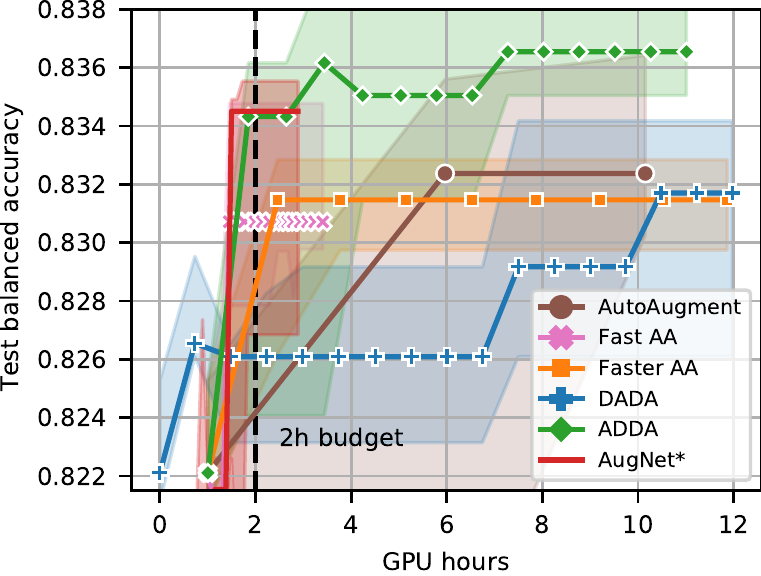}
        \caption{}
        \label{fig:mass-bars-curves}
    \end{subfigure}
    \hfill
    \begin{subfigure}[b]{0.49\linewidth}
        \centering
        \includegraphics[width=\linewidth]{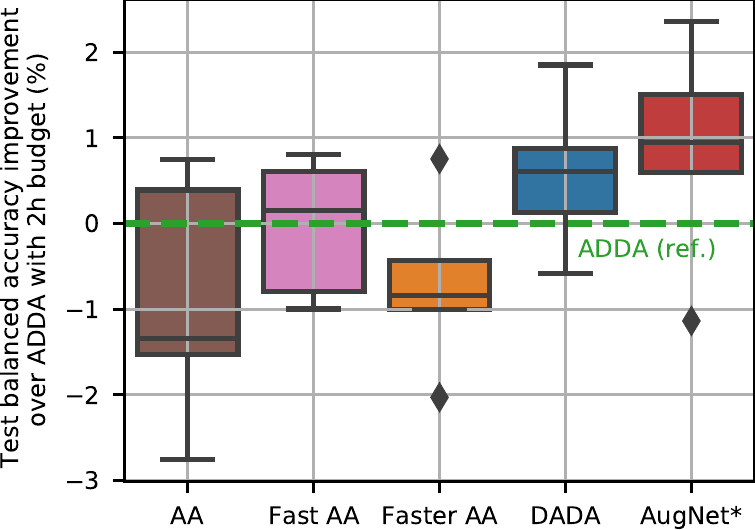}
        \vspace{\fill}
        \caption{}
        \label{fig:mass-bars-boxplots}
    \end{subfigure}
    \centering
    \caption{
    \textbf{(a)} Error bars correspond to 75\% confidence intervals over folds.
    \textbf{(b)} Fold-wise test balanced accuracy improvements with relation to ADDA after 2h of training. Performances at 2h of training were linearly interpolated from figure (a).
    }
    \label{fig:mass-bars}
\end{figure}

As explained in \autoref{sec:mass}, the plots from \autoref{fig:mass-benchmark} were obtained by computing median scores over 5 runs using different splits of the MASS dataset with a cross-validation scheme. In each run, the training, validation and test sets all contained data from different subjects. The same plot is presented in \autoref{fig:mass-bars-curves} with 75\% confidence error bars. Because of the well-known large inter-subject variability inherent to EEG recordings, we see that the between-splits variance is sometimes larger than the median performance gaps between methods, making it difficult to draw strong conclusions. In an effort to circumvent the inter-subject variability issue, relative scores were computed with regard to the ADDA method after a budget of 2 hours of training and plotted in \autoref{fig:mass-bars-boxplots}. Because we are computing performance gaps independently for each split, the latter are not hidden by the between-split variance in this case. This plot shows that in the context of a small training budget, AugNet delivers better performances in sleep stage classification compared to ADDA in four out of five runs.

\end{document}